\documentclass[11pt,a4paper]{article}
\usepackage[hyperref]{_ACL2021/acl2021}
\usepackage{times}
\usepackage{latexsym}

\usepackage{microtype}

\aclfinalcopy 
\def\aclpaperid{828} 

\usepackage{ownstyles}

\usepackage{booktabs}
\usepackage{mathtools}
\usepackage{multirow}
\usepackage{makecell}
\usepackage{array}
\usepackage{subcaption}
\usepackage{tabularx,colortbl}

\newcommand{\eg}{e.g.\xspace}
\newcommand{\ie}{i.e.\xspace}

\newcommand{\mybox}[2]{{\color{#1}\fbox{\normalcolor#2}}}
\newcolumntype{M}[1]{>{\centering\arraybackslash}m{#1}}

\newcommand{\bertBase}{BERT$_{\text{\ensuremath{\mathsf{base}}}}$\xspace}
\newcommand{\bertLarge}{BERT$_{\text{\ensuremath{\mathsf{large}}}}$\xspace}
\newcommand{\robertaBase}{RoBERTa$_{\text{\ensuremath{\mathsf{base}}}}$\xspace}
\newcommand{\robertaLarge}{RoBERTa$_{\text{\ensuremath{\mathsf{large}}}}$\xspace}
\newcommand{\albertBase}{ALBERT$_{\text{\ensuremath{\mathsf{base}}}}$\xspace}

\newcommand{\subsec}[1]{\noindent{\textbf{#1~~}}}
\newcommand{\devr}[0]{\texttt{dev-r}\xspace}
\newcommand{\devs}[0]{\texttt{dev-s}\xspace}

\newcommand{\class}[1]{{\colorbox{gray!7}{``#1''}}}

\newcommand{\hl}[1]{\mybox{magenta}{\strut#1}}
\definecolor{myGreen}{rgb}{0.04,0.58,0.22}
\definecolor{myBlue}{rgb}{0.1,0.4,0.56}

\newif\ifcomments

\commentstrue

\ifcomments
\newcommand{\comments}[1]{#1}
\else
\newcommand{\comments}[1]{}
\fi

\title{Out of Order: {H}ow Important Is The Sequential Order of Words \\ in a Sentence in {N}atural {L}anguage {U}nderstanding Tasks?}

\author{Thang M. Pham$^1$ \\
  {\small\texttt{thangpham@auburn.edu}} \\\And
  Trung Bui$^2$ \\
  {\small\texttt{bui@adobe.com}} \\\And
  Long Mai$^2$ \\
  {\small\texttt{malong@adobe.com}} \\\AND
  Anh Nguyen$^1$ \\
  \small{\texttt{anh.ng8@gmail.com}} \\\AND
    {\normalfont $^1$Auburn University~~~~~$^2$Adobe Research}
  }

\begin{document}

\maketitle

\begin{abstract}
Do state-of-the-art natural language understanding models care about word order?
Not always! 
We found 75\% to 90\% of the correct predictions of BERT-based classifiers, trained on many GLUE tasks, remain constant after input words are randomly shuffled.
Although BERT embeddings are famously contextual, the contribution of each individual word to classification is almost unchanged even after its surrounding words are shuffled.
BERT-based models exploit superficial cues (e.g. the sentiment of keywords in sentiment analysis; or the word-wise similarity between sequence-pair inputs in natural language inference) to make correct decisions when tokens are randomly shuffled.
Encouraging models to capture word order information improves the performance on most GLUE tasks and SQuAD 2.0.
Our work suggests that many GLUE tasks are not challenging machines to understand the meaning of a sentence.
\end{abstract}

\section{Introduction}

\begin{figure}[hbt!]
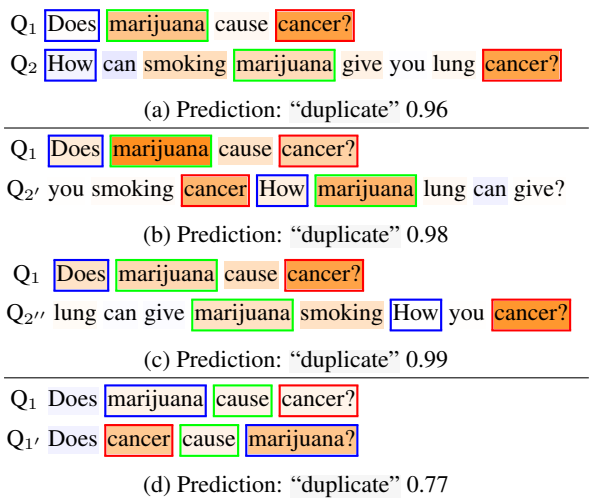

\centering
\setlength\tabcolsep{1.0pt}

\begin{subfigure}{\columnwidth}
\centering
\begin{tabular}{cp{0.92\linewidth}}
\footnotesize Q$_{1}$ & \footnotesize \text{\mybox{blue}{\colorbox{orange!1.24}{\strut Does}}} \mybox{green}{\colorbox{orange!46.317}{\strut marijuana}} \colorbox{orange!5.12}{\strut cause} \mybox{red}{\colorbox{orange!73.74199999999999}{\strut cancer?}} \\
\footnotesize Q$_{2}$ & \footnotesize \text{\mybox{blue}{\colorbox{blue!4.32}{\strut How}}} \colorbox{blue!8.871}{\strut can} \colorbox{orange!26.172}{\strut smoking} \mybox{green}{\colorbox{orange!27.87}{\strut marijuana}} \colorbox{orange!10.297}{\strut give} \colorbox{blue!0.9690000000000001}{\strut you} \colorbox{orange!5.593}{\strut lung} \mybox{red}{\colorbox{orange!71.691}{\strut cancer?}} \\
\end{tabular}
\caption{Prediction: \class{duplicate} 0.96}
\label{fig:teaser_a}
\end{subfigure}

\begin{subfigure}{\columnwidth}
\centering
\begin{tabular}{cp{0.92\linewidth}c}
\hline
\footnotesize Q$_{1}$ & \footnotesize \text{\mybox{blue}{\colorbox{orange!15.245000000000001}{\strut Does}}} \mybox{green}{\colorbox{orange!87.78399999999999}{\strut marijuana}} \colorbox{orange!21.18}{\strut cause} \mybox{red}{\colorbox{orange!35.46}{\strut cancer?}} \\
\footnotesize Q$_{2'}$ & \footnotesize \colorbox{orange!0.607}{\strut you} \colorbox{orange!2.779}{\strut smoking} \mybox{red}{\colorbox{orange!56.918}{\strut cancer}} \mybox{blue}{\colorbox{orange!11.154}{\strut How}} \mybox{green}{\colorbox{orange!57.884}{\strut marijuana}} \colorbox{orange!4.422000000000001}{\strut lung} \colorbox{blue!5.427}{\strut can} \colorbox{orange!1.646}{\strut give?} \\
\end{tabular}
\caption{Prediction: \class{duplicate} 0.98}
\label{fig:teaser_b}
\end{subfigure}

\begin{subfigure}{\columnwidth}
\centering
\begin{tabular}{cp{0.92\linewidth}c}

\footnotesize Q$_{1}$ & \footnotesize \text{\mybox{blue}{\colorbox{orange!22.078999999999997}{\strut Does}}} \mybox{green}{\colorbox{orange!30.357}{\strut marijuana}} \colorbox{orange!24.989}{\strut cause} \mybox{red}{\colorbox{orange!72.03699999999999}{\strut cancer?}} \\
\footnotesize Q$_{2''}$ & \footnotesize \colorbox{orange!3.485}{\strut lung} \colorbox{blue!1.2510000000000001}{\strut can} \colorbox{blue!1.6740000000000002}{\strut give} \mybox{green}{\colorbox{orange!23.0}{\strut marijuana}} \colorbox{orange!22.817}{\strut smoking} \mybox{blue}{\colorbox{blue!0.17700000000000002}{\strut How}} \colorbox{orange!0.759}{\strut you} \mybox{red}{\colorbox{orange!81.768}{\strut cancer?}} \\
\end{tabular}
\caption{Prediction: \class{duplicate} 0.99}
\label{fig:teaser_c}
\end{subfigure}

\begin{subfigure}{\columnwidth}
\centering
\begin{tabular}{cp{0.92\linewidth}c}
\hline
\footnotesize Q$_{1}$ & \footnotesize \colorbox{blue!4.356}{\strut Does} \mybox{blue}{\colorbox{orange!8.434999999999999}{\strut marijuana}} \mybox{green}{\colorbox{orange!8.632}{\strut cause}} \mybox{red}{\colorbox{orange!8.187999999999999}{\strut cancer?}} \\
\footnotesize Q$_{1'}$ & \footnotesize \colorbox{blue!5.325}{\strut Does} \mybox{red}{\colorbox{orange!42.734}{\strut cancer}} \mybox{green}{\colorbox{orange!9.112}{\strut cause}} \mybox{blue}{\colorbox{orange!44.756}{\strut marijuana?}} \\
\end{tabular}  
\caption{Prediction: \class{duplicate} 0.77}
\label{fig:teaser_d}
\end{subfigure}
\caption{
A RoBERTa-based model achieving a 91.12\% accuracy on QQP, here, correctly labeled a pair of Quora questions \class{duplicate} (a).
Interestingly, the predictions remain unchanged when all words in question Q$_{2}$ is randomly shuffled (b--c).
QQP models also often incorrectly label a real sentence and its shuffled version to be \class{duplicate} (d).
We found evidence that GLUE models rely heavily on words to make decisions \eg here, ``marijuana'' and ``cancer'' (more important words are \colorbox{orange!44.756}{\strut highlighted} by LIME).
Also, there exist self-attention matrices tasked explicitly with extracting word-correspondence between two input sentences regardless of the position of those words.
Here, the top-3 pairs of words assigned the highest self-attention weights at (layer 0, head 7) are inside \textcolor{red}{red}, \textcolor{green}{green}, and \textcolor{blue}{blue} rectangles, respectively.
}
\label{fig:teaser}
\end{figure}


Machine learning (ML) models recently achieved excellent performance on state-of-the-art benchmarks for evaluating natural language understanding (NLU).
In July 2019, RoBERTa \citep{liu2019roberta} was the first to surpass a human baseline on GLUE \citep{wang2018glue}.
Since then, 13 more methods have also outperformed humans on the \citealt{glueleaderboard}.
Notably, at least 8 out of the 14 solutions are based on BERT \citep{devlin2019bert}---a transformer architecture that learns representations via a bidirectional encoder.
Given their superhuman GLUE-scores, how do BERT-based models solve NLU tasks? How do their NLU capability differs from that of humans?


We shed light into these important questions by examining model sensitivity to the order of words.
Word order is one of the key characteristics of a sequence and is tightly constrained by many linguistic factors including syntactic structures, subcategorization, and discourse \citep{elman1990finding}.
Thus, arranging a set of words in a correct order is considered a key problem in language modeling~\citep{hasler2017comparison,zhang2015discriminative}.

Therefore, a natural question is: \textbf{Do BERT-based models trained on GLUE care about the order of words in a sentence?}
\citet{lin2019open} found that pretrained BERT captures word-order information in the first three layers.
However, it is unknown whether BERT-based classifiers actually use word order information when performing NLU tasks.
Recently, \citet{wang2020structbert} showed that incorporating additional word-ordering and sentence-ordering objectives into BERT pretraining could lead to text representations (StructBERT) that enabled improved GLUE scores.
However, StructBERT findings are inconclusive across different GLUE tasks and models.
For example, in textual entailment \citep[RTE]{wang2018glue}, StructBERT improved the performance for \bertLarge but hurt the performance for RoBERTa (Table~\ref{table:miniQ1}d).

\citealt{wang2020structbert} motivated interesting questions: \textbf{Are state-of-the-art BERT-based models using word order information when solving NLU tasks?
If not, what cues do they rely on?}
To the best of our knowledge, our work is the first to study the above questions for an NLU benchmark (GLUE).
We tested BERT-, RoBERTa-, and ALBERT-based \citep{lan2020albert} models on 7 GLUE tasks where the words of only one select sentence in the input text are shuffled at varying degrees.
An ideal agent that truly understands language is expected to choose a \class{reject} option when asked to classify a sentence whose words are randomly shuffled.
Alternatively, given shuffled input words, true NLU agents are expected to perform at random chance in multi-way classification that has no \class{reject} options (Fig.~\ref{fig:teaser_b}).
Our findings include:

\begin{enumerate}
    \item 
    65\% of the groundtruth labels of 5 GLUE tasks can be predicted when the words in one sentence in each example are shuffled (Sec.~\ref{sec:wos}).
    
    \item Although pretrained BERT embeddings are known to be contextual, in some GLUE tasks, the contribution of an individual word to classification is almost unchanged even after its surrounding words are shuffled (Sec.~\ref{sec:heatmap_similarity}).
    
    \item In sentiment analysis (SST-2), the polarity of a single salient word is $\geq$ 60\% predictive of an entire sentence's label (Sec.~\ref{sec:sst2}).
    
    \item BERT-based models trained on sequence-pair GLUE tasks used a set of self-attention heads for finding similar tokens shared between the two inputs (Sec.~\ref{sec:other_cues}).

    \item Encouraging RoBERTa-based models to be more sensitive to word order improves the performance on SQuAD 2.0 and most GLUE tasks tested (\ie except for SST-2) (Sec.~\ref{sec:extra_finetuning}).
\end{enumerate}

Despite their superhuman scores, most GLUE-trained models behave similarly to Bag-of-Words (BOW) models, which are prone to naive mistakes (Fig.~\ref{fig:teaser}b--d).
Our results also suggest that GLUE does not necessarily require syntactic information or complex reasoning.

\section{Methods}
\label{sec:methods}

\subsection{Datasets}

We chose GLUE because of three reasons: 
(1) GLUE is a common benchmark for NLU evaluation \citep{wang2018glue};
(2) there exist NLU models (\eg RoBERTa) that outperform humans on GLUE, making an important case for studying their behaviors;
(3) it is unknown how sensitive GLUE-trained models are to word order and whether GLUE requires them to be sensitive \citep{wang2020structbert}.

\subsec{Tasks} 
Out of 9 GLUE tasks, we chose all 6 binary-classification tasks because they share the same random baseline of 50\% accuracy and enable us to compare models’ word-order sensitivity across tasks.
Six tasks vary from acceptability (CoLA \citealt{warstadt2018neural}), to natural language inference (QNLI \citealt{rajpurkar2016squad}), RTE \citep{wang2018glue}, paraphrase (MRPC \citealt{dolan2005automatically}, QQP \citealt{quora2020dataset}), and sentiment analysis (SST-2 \citealt{socher2013recursive}).

We also performed our tests on STS-B \citep{cer2017semeval}---a regression task of predicting the semantic similarity of two sentences.\footnote{We did not choose WNLI \citep{levesque2012winograd} as model performance is not substantially above random baseline.}
While CoLA and SST-2 require single-sentence inputs, all other tasks require sequence-pair inputs.

\subsec{Reject options}
For all binary-classification tasks (except SST-2), the {negative} label is considered the reject option (\eg QQP models can choose \class{not duplicate} in Fig.~\ref{fig:teaser_b} to reject shuffled inputs).

\subsec{Metrics} 
We use accuracy scores to evaluate the binary classifiers (for ease of interpretation) and Spearman correlation to evaluate STS-B regressors, following \citet{wang2018glue}.

\subsection{Classifiers}
\label{sec:classifiers}

We tested BERT-based models because (1) they outperformed humans on the \citealt{glueleaderboard}; and (2) the pretrained BERT was shown to capture word positional information \citep{lin2019open}.

\subsec{Pretrained BERT encoders}
We tested three sets of classifiers finetuned from three different, pretrained BERT variants: BERT, RoBERTa, and ALBERT, downloaded from \citet{huggingface2020pretrained}.
The pretrained models are the ``base'' versions \ie bidirectional transformers with 12 layers and 12 self-attention heads.
The pretraining corpus varies from uncased (BERT, ALBERT) to case-sensitive English (RoBERTa).


\subsec{Classifiers}
For each of the seven GLUE tasks, we added one classification layer on top of each of the three pretrained BERT encoders and finetuned the entire model.
Unless otherwise noted, the mean performance per GLUE task was averaged over three classifiers.
Each model's performance matches either those reported on \citet{huggingface2020pretrained} or the original papers (Table~\ref{table:original_performance}).


\subsec{Hyperparameters} 
Following \citet{devlin2019bert}, we finetuned classifiers for 3 epochs using Adam \citep{kingma2014adam} with a learning rate of 0.00002, $\beta_1$ = 0.9, $\beta_2$ = 0.999, $\epsilon = 10^{-8}$. 
We used a batch size of 32, a max sequence length of 128, and dropout on all layers with a probability of 0.1.





\subsection{Constructing sets of real and shuffled examples for experiments}
\label{sec:shuffling_methods}

\subsec{Modifying one sentence} As GLUE tasks vary in the number of inputs (one or two input sequences) and the sequence type per input (a sentence or a paragraph), we only re-ordered the words in one \emph{sentence} from only one input while keeping the rest of the inputs unchanged. 
Constraining the modifications to a single sentence enables us to measure (1) the importance of word order in a single sentence; and (2) the interaction between the shuffled words and the unchanged, real context.

\subsec{Random shuffling methods} To understand model behaviors across varying degrees of word-order distortions, we experimented with three tests: randomly shuffling n-grams where n = $\{1,2,3\}$.

Shuffling 1-grams is a common technique for analyzing word-order sensitivity \cite{sankar-etal-2019-neural,zanzotto2020kermit}.
We split a given sentence by whitespace into a list of n-grams, and re-combined them, in a random order, back into a ``shuffled'' sentence (see Table~\ref{fig:example_shuffling_types} for examples).
The ending punctuation was kept intact.
We re-sampled a new random permutation until the shuffled sentence was different from the original sentence.


\begin{table}[ht]
\centering
\setlength\tabcolsep{1pt}
\centering
\begin{tabular}{lp{0.92\linewidth}}
%
\footnotesize How can smoking marijuana give you lung cancer? \\ 
\footnotesize Q$_3$ \footnotesize \hl{lung cancer} \hl{marijuana give you} \hl{How can smoking}? \\ 
\footnotesize Q$_2$ \footnotesize \hl{smoking marijuana}~ \hl{lung cancer} \hl{give you} \hl{How can}? \\ 
\footnotesize Q$_1$ \footnotesize \hl{marijuana} \hl{can} \hl{cancer} \hl{How} \hl{you} \hl{smoking} \hl{give} \hl{lung}? \\ 
\footnotesize Q$_s$ How can smoking \hl{cancer} give you lung \hl{marijuana}? \\ 
\end{tabular}

\caption{A real question on Quora (QQP dataset) and its three modified versions (Q$_3$ to Q$_1$) created by randomly shuffling 3-grams, 2-grams, and 1-grams, respectively.
Q$_s$ was created by swapping two random nouns.
}
\label{fig:example_shuffling_types}
\end{table}


As the label distributions, dev-set sizes, and the performance of models vary across GLUE tasks, to compare word-order sensitivity across tasks, we tested each model on two sets: (1) \devr \ie a subset of the original dev-set (Sec.~\ref{sec:dataset_construction}); and (2) \devs \ie a clone of version of \devr but that each example has one sentence with re-ordered words (Sec.~\ref{sec:shuffled_sets}).


\subsubsection{Selecting real examples}
\label{sec:dataset_construction}

For each pair of (task, classifier), we selected a subset of dev-set examples via the following steps:

\begin{enumerate}
    \item For tasks with either a single-sequence or a sequence-pair input, we used examples where the input sequence to be modified has only one sentence\footnote{We used NLTK sentence splitter \citep{BirdKleinLoper09} to detect text that has more than one sentence.} that has more than 3 tokens (for shuffling 3-grams to produce a sentence different from the original sentence).
    \item We only selected the examples that were correctly classified by the classifier (to study what features were important for high accuracy).
    \item We balanced the numbers of positive and negative examples by removing random examples from the larger-sized class. 
\end{enumerate}

That is, on average, we filtered out $\sim$34\% of the original data.
See Table~\ref{table:preprocessing_stats} for the total number of examples remaining after each filtering step above.

\subsubsection{Creating shuffled sets}
\label{sec:shuffled_sets}

For each task, we cloned the \devr sets above and modified each example to create a ``shuffled'' set (a.k.a. \devs) per shuffling method.

Specifically, a CoLA and SST-2 example contains only a single sentence and we modified that sentence.
Each QQP, MRPC and STS-B example has two sentences and we modified the first sentence.
An RTE example has a pair of (premise, hypothesis), and we modified the hypothesis since it is a single sentence while premises are paragraphs.
Each QNLI example contains a pair of (question, answer) and we modified the question, which is a sentence, while an answer is often a paragraph.

\section{Experiments and Results}


\subsection{How much is word order information required for solving GLUE tasks?}
\label{sec:wos}

GLUE has been a common benchmark for evaluating NLU progress.
But, do GLUE tasks require models to use word order and syntactic information?
We shed light into this question by testing model performance when word order is increasingly randomized.

If a task strictly requires words to form a semantically meaningful sentence, then randomly re-positioning words in correctly-classified sentences will cause model accuracy to drop from 100\% to 50\% (\ie the random baseline $b$ for binary-classification tasks with two balanced classes).
Thus, to compare model-sensitivity across tasks, we use a Word-Order Sensitivity score (WOS):

\begin{equation}
s = (100 - p) / (100 - b)
\end{equation}

where $p \in [50, 100]$ is the accuracy of a GLUE-trained model evaluated on a $\devs$ set (described in Sec.~\ref{sec:shuffled_sets}) and $s \in [ 0, 1 ]$.
Here, $b = 50$.

\paragraph{Experiments}
For each GLUE task, we computed the mean accuracy and confidence score over three classifiers (BERT, RoBERTa, and ALBERT-based) on \devs sets created by shuffling 1-grams, 2-grams, and 3-grams.
The results reported in Table~\ref{table:miniQ1} were averaged over 10 random shuffling runs (\ie 10 random seeds) per n-gram type, and then averaged over 3 models per task.

\paragraph{Results}

We found that for CoLA, \ie detecting grammatically incorrect sentences, the model accuracy, on average, drops to near random chance \ie between 50.69\% and 56.36\% (Table~\ref{table:miniQ1}b) when n-grams are shuffled.
That is, most of examples were classified into \class{unacceptable} after n-gram shuffling, yielding $\sim$50\% accuracy (see Fig.~\ref{fig:qualitative_examples_cola} for qualitative examples).

Surprisingly, for the rest of the 5 out of 6 binary-classification tasks (\ie except CoLA), between 75\%  and 90\% of the originally correct predictions remain constant after 1-grams are randomly re-ordered (Table~\ref{table:miniQ1}b; 1-gram).
These numbers increase as the shuffled n-grams are longer (\ie as n increases from 1$\to$3), up to 95.32\% (Table~\ref{table:miniQ1}b; QNLI).
Importantly, given an average dev-set accuracy of 86.35\% for these 5 tasks, \textbf{at least} 86.35\% $\times$ 75\% $\approx$ \textbf{65\% of the groundtruth labels of these 5 GLUE tasks can be predicted when all input words in one sentence are randomly shuffled.}

Additionally, on average over three n-gram types, models trained on these five GLUE tasks are from 2 to 10 times more \emph{insensitive} to word-order randomization than CoLA models (Table~\ref{table:miniQ1}c).
That is, if not explicitly tasked with checking for grammatical errors, GLUE models mostly will not care about the order of words in a sentence (see qualitative examples in Figs.~\ref{fig:teaser}, \ref{fig:qualitative_examples_cola}--\ref{fig:qualitative_examples_rte}).
Consistently, the confidence scores of BERT-based models for five non-CoLA tasks only dropped $\sim$2\% when 1-grams are shuffled (Table~\ref{table:miniQ1}).

Consistently across three different BERT ``base" variants and a RoBERTa ``large" model (Table \ref{table:shuffle}), our results suggest that word order and syntax, in general, are not necessarily required to solve GLUE.

\subsec{2-noun swaps}
Besides shuffled n-grams, we also repeated all experiments with more syntactically-correct modified inputs where only two random nouns in a sentence were swapped (Table~\ref{fig:example_shuffling_types}; Q$_s$).
This is a harder test for NLU models since the meaning of a sentence with two nouns swapped often changes while its syntax remains correct.
We found the conclusions to generalize to this setting. 
That is, the models hardly changed predictions although the meanings of the original sentence and its swapped version are different (Table~\ref{table:miniQ1}b; 2-noun swap vs. 1-gram).





\subsection{How sensitive are models trained to predict the similarity of two sentences?}
\label{sec:qqp_stsb}

An interesting hypothesis is that models trained explicitly to evaluate the semantic similarity of two sentences should be able to tell apart real from shuffled examples.
Intuitively, word order information is essential for understanding what an entire sentence means and, therefore, for predicting whether two sentences convey the same meaning.

We tested this hypothesis by analyzing the sensitivity of models trained on QQP and STS-B---two prominent GLUE tasks for predicting semantic similarity of a sentence pair.
While QQP is a binary classification task, STS-B is a regression task where a pair of two sentences is given a score $\in [0,5]$ denoting their semantic similarity. 

\paragraph{Experiments} 
We tested the models on \devr and \devs sets (see Sec.~\ref{sec:shuffled_sets}) where in each pair, the word order of the first sentence was randomized while the second sentence was kept intact.

\paragraph{QQP results} 
Above 83\% of QQP models' correct predictions on real pairs remained unchanged after word-order randomization (see Figs.~\ref{fig:teaser_a}--c for examples).

\begin{figure}[ht]
\centering
\includegraphics[width=\linewidth]{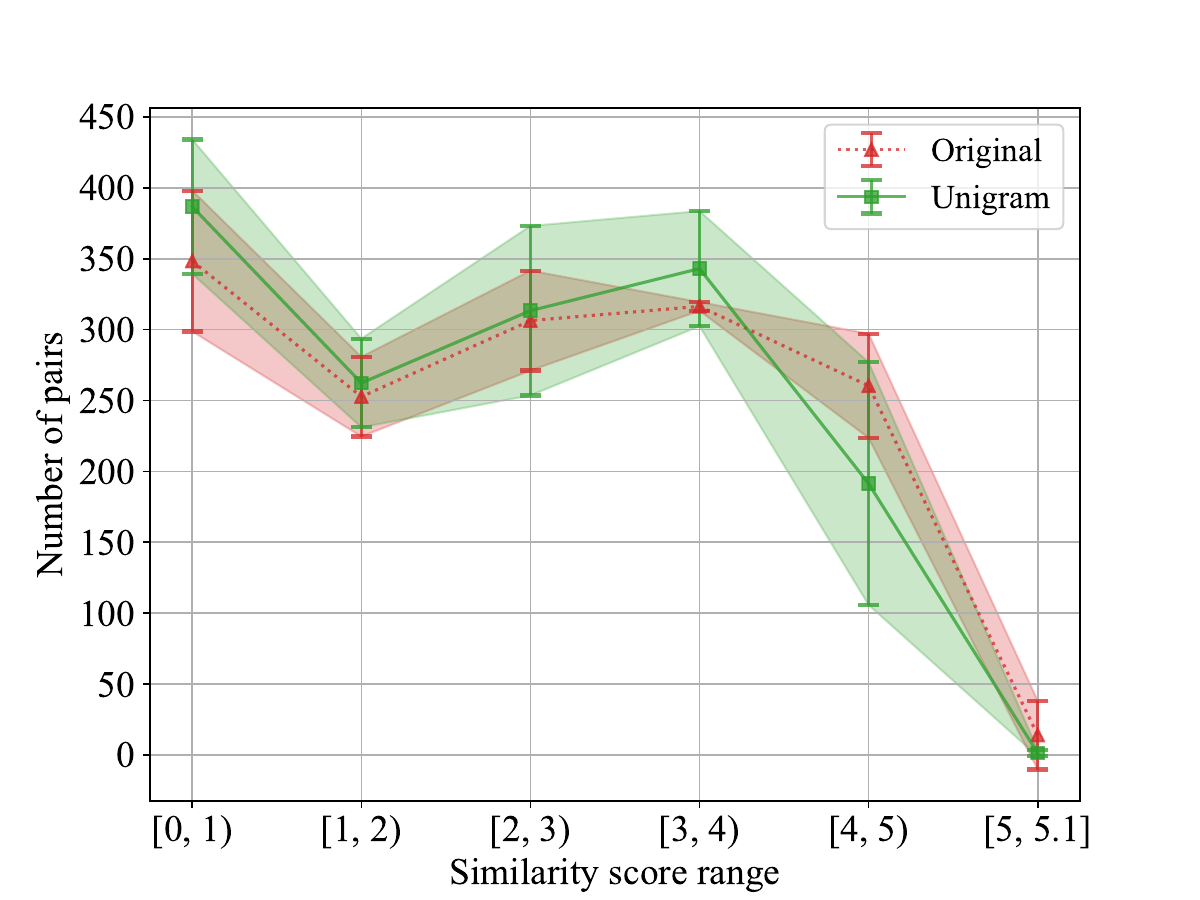}
\caption{
The distribution of similarity scores over 6 ranges for the (real, shuffled) pairs in \devs (\textcolor{myGreen}{green}) is highly similar to that for (real, real) STS-B pairs in \devr (\textcolor{red}{red}).
The statistics in each range were computed over 3 models (BERT, RoBERTa, and ALBERT).
}
\label{fig:stsb_ori_shuffle}

\end{figure}

\paragraph{STS-B results} Similarly, STS-B model performance only drops marginally, \ie less than 2 points from 89.67 to 87.80 in Spearman correlation (Table~\ref{table:miniQ1}; STS-B).
Since a STS-B model outputs a score $\in [0,5]$, we binned the scores into 6 ranges.
One might expect STS-B models to assign near-zero similarity scores to most modified pairs.
However, the distributions of similarity scores for the modified and real pairs still closely match up (Fig.~\ref{fig:stsb_ori_shuffle}).
In sum, \textbf{despite being trained explicitly on predicting semantic similarity of sentence pairs, QQP and STS-B are surprisingly insensitive to n-gram shuffling, exhibiting naive understanding of sentence meanings}.

\begin{table*}[t]
\centering\small
\setlength\tabcolsep{1.5pt}
\begin{tabular}{lcccccccccccccc}
\toprule
Task & \multicolumn{2}{c}{(a) Perf. on \devr} & \multicolumn{4}{c}{(b) Performance on \devs} & \multicolumn{3}{c}{(c) Word-Order Sensitivity} & \multicolumn{3}{c}{(d) StructBERT improvements} \\
\cmidrule(l{1pt}r{1pt}){2-3} \cmidrule(l{1pt}r{1pt}){4-7} \cmidrule(l{1pt}r{1pt}){8-10} \cmidrule(l{1pt}r{1pt}){11-13}
& Models & Baseline & 2-noun swap & 1-gram & 2-gram & 3-gram & 1-gram & 2-gram & 3-gram & \bertBase & \bertLarge & RoBERTa \\

\toprule 
\texttt{CoLA}  & 100 & 50 & 71.75 & 50.69 & 53.98 & 56.36 & \textbf{0.99} & \textbf{0.92} & \textbf{0.87} & \textbf{+4.9} & +4.8 & +1.4 \\
& (0.93) &  & (0.91) & (0.95) & (0.94) & (0.92) &  &  &  &  &  &  \\
\cmidrule{1-13}
\texttt{RTE}  & 100 & 50 & 85.86 & 75.69 & 81.89 & 85.18 & 0.49 & 0.36 & 0.30 & N/A & \textbf{+13.0} & --0.9 \\
& (0.81) & & (0.81) & (0.80) & (0.80) & (0.79) &  &  &  &  &  &  \\
\cmidrule{1-13}
\texttt{QQP}  & 100 & 50 & 86.90 & 83.19 & 88.02 & 89.04 & 0.34 & 0.24 & 0.22 & +0.7 & +1.2 & +0.5 \\
& (0.98) &  & (0.96) & (0.96) & (0.96) & (0.96) &  &  &  &  &  &  \\
\cmidrule{1-13}
\texttt{MRPC}  & 100 & 50 & 96.51 & 83.89 & 87.1 & 89.38 & 0.32 & 0.26 & 0.21 & N/A & +3.9 & \textbf{+1.7} \\
& (0.91) &  & (0.91) & (0.89) & (0.90) & (0.90) &  &  &  &  &  &  \\
\cmidrule{1-13}
\texttt{SST-2}  & 100 & 50 & 97.78 & 84.04 & 88.35 & 90.56 & 0.32 & 0.23 & 0.19 & +0.2 & +0.3 & +0.4 \\
& (0.99) &  & (0.98) & (0.96) & (0.97) & (0.97) &  &  &  &  &  &  \\
\cmidrule{1-13}
\texttt{QNLI} & 100 & 50 & 94.31 & 89.42 & 93.85 & 95.32 & 0.21 & 0.12 & 0.09 & N/A & +3.0 & +0.3 \\
& (0.98) &  & (0.97) & (0.96) & (0.97) & (0.98) &  &  &  &  &  &  \\
\hline
\cmidrule{1-13}
\texttt{STS-B} & 89.67 & N/A & 88.93 & 87.80 & 88.66 & 88.95 & N/A & N/A & N/A & N/A & N/A & N/A \\
\bottomrule

\end{tabular}
\caption{
All results (a--c) are reported on the GLUE \devr sets \ie 100\% accuracy (a).
Shuffling n-grams caused the accuracy to drop (b) the largest for CoLA and the least for QNLI.
Each row is computed by averaging the results of 3 BERT-based models and 10 random shuffles.
From top to bottom, the Word-Order Sensitivity (WOS) is sorted descendingly (c) and is consistent across three types of n-grams \ie WOS scores decrease from top down and from left to right.
In contrast, the StructBERT results (d), taken from Table 1 and 4 in \citealt{wang2020structbert}, showed inconsistent improvements across different tasks.
STS-B results are in scaled Spearman correlation.
In addition to small accuracy drops, the mean confidence scores of all classifiers---reported in parentheses e.g. ``(0.93)''---also changed marginally after words are shuffled (a vs. b).
}
\label{table:miniQ1}
\end{table*}

\subsection{How important are words to classification after their context is shuffled?}
\label{sec:heatmap_similarity}

BERT representations for tokens are known to be highly contextual \cite{ethayarajh2019contextual}.
However, after finetuning on GLUE, would the importance of a word to classification drop after its context is shuffled?

%

To answer the above question, we used LIME \cite{ribeiro2016should} to compute word importance.
\subsec{LIME} computes a score $\in [-1, 1]$ for each token in the input denoting how much its presence contributes for or against the network's predicted label (Fig.~\ref{fig:teaser}; \colorbox{orange!44.756}{\strut highlights}).
The importance score per word $w$ is intuitively the mean confidence-score drop over a set of randomly-masked versions of the input when $w$ is masked out.

\paragraph{Experiments}
We chose to study RoBERTa-based classifiers here because they have the highest GLUE scores among the three BERT variants considered.
We observed that \textbf{62.5\% (RTE) to 79.6\% (QNLI) of the \devr examples were consistently, correctly classified into the same labels in all 5 different random shuffles} (\ie 5 different random seeds).
We randomly sampled 100 such examples per binary-classification task and computed their LIME attribution maps to compare the similarity between the LIME heatmaps before and after unigrams are randomly misplaced.

%

\subsec{Results}
On CoLA and RTE, the importance of words (\ie mean absolute value of LIME-attribution per word), \emph{decreased} substantially by 0.036 and 0.019, respectively.
That is, the individual words become \emph{less important} after their context is distorted---a behavior expected when CoLA and RTE have the highest WOS scores (Table~\ref{table:miniQ1}).
In contrast, for the other 4 tasks, word importance only changed marginally (by 0.008, \ie 4.5$\times$ smaller than the 0.036 change in CoLA).
That is, \textbf{except for CoLA and RTE models, the contribution of a word to classification is almost unchanged even after the context of each word is randomly shuffled} (Fig.~\ref{fig:teaser_a}--c).
This result suggests that the word embeddings after finetuning on GLUE became much less contextual than the pretrained BERT embeddings \cite{ethayarajh2019contextual}.

\subsection{If not word order, then what do classifiers rely on to make correct predictions?}
\label{sec:other_cues}

Given that all non-CoLA models are highly insensitive to word-order randomization, how did they arrive at correct decisions when words are shuffled?

We chose to answer this question for SST-2 and QNLI because they have the lowest WOS scores across all 6 GLUE tasks tested (Table~\ref{table:miniQ1}) and they are representative of single-sentence and sequence-pair tasks, respectively.

\subsubsection{SST-2: Salient words are highly predictive of sentence labels}
\label{sec:sst2}

As 84.04\% of the SST-2 correct predictions did not change after word-shuffling (Table~\ref{table:miniQ1}b), a common hypothesis is that the models might rely heavily on a few key words to classify an entire sentence.

\begin{figure}[ht]
\centering\small
\setlength\tabcolsep{2pt}
\begin{tabular}{|l|l|c|}
\hline

S & \footnotesize \colorbox{blue!0.698}{\strut the} \colorbox{orange!0.545}{\strut film} \colorbox{orange!12.777}{\strut 's} \colorbox{orange!14.689}{\strut performances} \colorbox{blue!7.579}{\strut are} \colorbox{orange!33.821}{\strut thrilling} \colorbox{orange!0}{\strut .} & 1.00 \\
\hline
\hline
S$_{1}$ & \footnotesize \colorbox{blue!1.8739999999999999}{\strut the} \colorbox{orange!6.063000000000001}{\strut film} \colorbox{orange!31.45}{\strut thrilling} \colorbox{orange!2.629}{\strut performances} \colorbox{blue!3.9350000000000005}{\strut are} \colorbox{orange!12.623999999999999}{\strut 's} \colorbox{orange!0}{\strut .} & 1.00 \\
\hline
S$_{2}$ & \footnotesize \colorbox{orange!3.0620000000000003}{\strut 's} \colorbox{orange!22.599}{\strut thrilling} \colorbox{orange!4.453}{\strut film} \colorbox{blue!0.126}{\strut are} \colorbox{orange!11.219}{\strut performances} \colorbox{orange!5.185}{\strut the} \colorbox{orange!0}{\strut .} & 1.00 \\
\hline
S$_{3}$ & \footnotesize \colorbox{blue!1.113}{\strut 's} \colorbox{orange!28.643}{\strut thrilling} \colorbox{blue!3.9079999999999995}{\strut are} \colorbox{orange!5.13}{\strut the} \colorbox{orange!7.149}{\strut performances} \colorbox{blue!1.425}{\strut film} \colorbox{orange!0}{\strut .} & 1.00 \\
\hline
\end{tabular}
\caption{
An original SST-2 dev-set example (S) and its three shuffled versions (S$_1$ to S$_3$) were all correctly labeled \class{positive} by a RoBERTa-based classifier with high confidence scores (right column).
}
\label{fig:sst2}
\end{figure}

\paragraph{Experiments}
To test this hypothesis, we took all SST-2 \devr examples whose all 5 randomly shuffled versions were all correctly labeled by a RoBERTa-based classifier (\ie this ``5/5'' subset is $\sim$65\% of the dev-set).
We used LIME to produce a heatmap of the importance of words in each example.

We identified the polarity of each top-1 most important word (\ie the highest LIME-attribution score) per example by looking it up in the Opinion Lexicon \cite{hu2004mining} of 2,006 positive and 4,783 negative words.
$\sim$57\% of these top-1 words were found in the dictionary and labeled either \class{positive} or \class{negative} (see Table~\ref{table:sst2_lexicons}).

\paragraph{Results}
We found that if the top-1 word has a positive meaning, then there is a 100\% probability that the sentence's label is \class{positive}.
For example, the word ``thrilling'' in a movie review indicates a \class{positive} sentence (see Fig.~\ref{fig:sst2}).
Similarly, the conditional probability of a sentence being labeled \class{negative} given a negative top-1 word is 94.4\%.
That is, given this statistics, the SST-2 label distribution and model accuracy, \textbf{at least 60\% of the SST-2 dev-set examples can be correctly predicted from only a single top-1 salient word.}


We also reached similar conclusions when experimenting with ALBERT classifiers and the SentiWords dictionary \cite{gatti2015sentiwords} (see Table~\ref{table:sst2_lexicons}).

\begin{figure*}[t]
\centering\small 
\setlength\tabcolsep{2pt}
\begin{tabular}{|c|p{0.85\linewidth}|c|}

\hline
\multicolumn{2}{|c|}{\multirow{2}{*}{QNLI sentence-pair inputs and their LIME attributions (\colorbox{blue!100}{\strut \textcolor{white}{negative -1}}, neutral 0, \colorbox{orange!100}{\strut positive +1})} } & \multirow{2}{*}{\makecell{Confidence \\ score}} \\
\multicolumn{2}{|c|}{\multirow{2}{*}{}} & \\
\hline

Q & \footnotesize \colorbox{orange!6.337}{\strut How} \colorbox{orange!41.776}{\strut long} \colorbox{orange!11.481}{\strut did} \mybox{green}{\colorbox{orange!14.180000000000001}{\strut Phillips}} \mybox{red}{\colorbox{orange!24.573999999999998}{\strut manage}} \colorbox{orange!1.6650000000000003}{\strut the} \mybox{blue}{\colorbox{orange!12.045}{\strut Apollo}} \colorbox{blue!20.738}{\strut missions?} & \multirow{2}{*}{1.00} \\
\cline{1-2}
A & \footnotesize \colorbox{blue!10.181}{\strut Mueller} \colorbox{orange!0.8099999999999999}{\strut agreed,} \colorbox{orange!1.4040000000000001}{\strut and} \mybox{green}{\colorbox{orange!14.111}{\strut Phillips}} \mybox{red}{\colorbox{orange!23.183999999999997}{\strut managed}} \mybox{blue}{\colorbox{blue!7.180000000000001}{\strut Apollo}} \colorbox{orange!20.158}{\strut from} \colorbox{orange!7.441000000000001}{\strut January} \colorbox{orange!16.778000000000002}{\strut 1964,} \colorbox{orange!27.162}{\strut until} \colorbox{orange!0.407}{\strut it} \colorbox{blue!7.6160000000000005}{\strut achieved} \colorbox{blue!1.994}{\strut the} \colorbox{blue!5.886}{\strut first} \colorbox{orange!2.646}{\strut manned} \colorbox{blue!6.69}{\strut landing} \colorbox{blue!1.472}{\strut in} \colorbox{orange!4.677}{\strut July} \colorbox{orange!3.386}{\strut 1969,} \colorbox{orange!2.0660000000000003}{\strut after} \colorbox{orange!0.7250000000000001}{\strut which} \colorbox{orange!0.5760000000000001}{\strut he} \colorbox{blue!1.214}{\strut returned} \colorbox{blue!2.827}{\strut to} \colorbox{blue!0.7979999999999999}{\strut Air} \colorbox{orange!0.909}{\strut Force} \colorbox{blue!5.378}{\strut duty.} &  \\
\hline\hline
Q$_{1}$ & \footnotesize \text{\mybox{blue}{\colorbox{orange!4.91}{\strut Apollo}}} \colorbox{blue!2.5860000000000003}{\strut the} \mybox{red}{\colorbox{orange!24.802}{\strut Phillips}} \colorbox{orange!2.393}{\strut How} \colorbox{blue!43.228}{\strut missions} \colorbox{orange!29.720000000000002}{\strut long} \colorbox{orange!29.012}{\strut did} \mybox{green}{\colorbox{orange!13.279}{\strut manage?}} & \multirow{2}{*}{0.96} \\
\cline{1-2}
A & \footnotesize \colorbox{blue!6.422999999999999}{\strut Mueller} \colorbox{blue!1.6039999999999999}{\strut agreed,} \colorbox{blue!3.817}{\strut and} \mybox{red}{\colorbox{orange!0.738}{\strut Phillips}} \mybox{green}{\colorbox{orange!56.81}{\strut managed}} \mybox{blue}{\colorbox{blue!6.753000000000001}{\strut Apollo}} \colorbox{orange!32.293}{\strut from} \colorbox{orange!8.291}{\strut January} \colorbox{orange!21.742}{\strut 1964,} \colorbox{orange!24.987000000000002}{\strut until} \colorbox{blue!2.573}{\strut it} \colorbox{blue!5.486}{\strut achieved} \colorbox{blue!7.025}{\strut the} \colorbox{blue!6.451}{\strut first} \colorbox{blue!2.718}{\strut manned} \colorbox{blue!10.646}{\strut landing} \colorbox{blue!1.532}{\strut in} \colorbox{orange!2.653}{\strut July} \colorbox{orange!1.369}{\strut 1969,} \colorbox{orange!3.2169999999999996}{\strut after} \colorbox{blue!2.603}{\strut which} \colorbox{blue!4.7509999999999994}{\strut he} \colorbox{blue!5.269}{\strut returned} \colorbox{orange!0.8009999999999999}{\strut to} \colorbox{blue!2.247}{\strut Air} \colorbox{blue!3.2329999999999997}{\strut Force} \colorbox{blue!2.965}{\strut duty.} &  \\
\hline\hline
Q$_{2}$ & \footnotesize \text{\mybox{green}{\colorbox{orange!28.458}{\strut Phillips}}} \colorbox{orange!26.735999999999997}{\strut long} \mybox{red}{\colorbox{orange!32.352}{\strut manage}} \colorbox{orange!8.652}{\strut How} \colorbox{blue!17.794999999999998}{\strut missions} \colorbox{orange!6.322}{\strut the} \mybox{blue}{\colorbox{orange!16.750999999999998}{\strut Apollo}} \colorbox{blue!17.713}{\strut did?} & \multirow{2}{*}{0.97} \\
\cline{1-2}
A & \footnotesize \colorbox{blue!11.193}{\strut Mueller} \colorbox{blue!3.982}{\strut agreed,} \colorbox{blue!3.6310000000000002}{\strut and} \mybox{green}{\colorbox{orange!17.187}{\strut Phillips}} \mybox{red}{\colorbox{orange!54.652}{\strut managed}} \mybox{blue}{\colorbox{blue!2.146}{\strut Apollo}} \colorbox{orange!16.169}{\strut from} \colorbox{orange!4.8660000000000005}{\strut January} \colorbox{orange!17.671999999999997}{\strut 1964,} \colorbox{orange!26.293}{\strut until} \colorbox{blue!1.171}{\strut it} \colorbox{orange!0.581}{\strut achieved} \colorbox{orange!0.652}{\strut the} \colorbox{blue!5.2459999999999996}{\strut first} \colorbox{orange!6.419}{\strut manned} \colorbox{blue!1.788}{\strut landing} \colorbox{blue!1.7690000000000001}{\strut in} \colorbox{blue!0.041}{\strut July} \colorbox{orange!5.67}{\strut 1969,} \colorbox{orange!1.7489999999999999}{\strut after} \colorbox{orange!0.382}{\strut which} \colorbox{blue!3.8699999999999997}{\strut he} \colorbox{blue!3.039}{\strut returned} \colorbox{blue!0.9440000000000001}{\strut to} \colorbox{orange!1.1860000000000002}{\strut Air} \colorbox{orange!1.078}{\strut Force} \colorbox{blue!4.7989999999999995}{\strut duty.} &  \\
\hline\hline
Q$_{s}$ & \footnotesize \colorbox{orange!6.132}{\strut How} \colorbox{orange!46.922999999999995}{\strut long} \colorbox{orange!13.132}{\strut did} \mybox{green}{\colorbox{orange!4.340999999999999}{\strut Apollo}} \mybox{red}{\colorbox{orange!30.375000000000004}{\strut manage}} \colorbox{blue!3.309}{\strut the} \mybox{blue}{\colorbox{orange!12.558}{\strut Phillips}} \colorbox{blue!25.404}{\strut missions?} & \multirow{2}{*}{0.99} \\
\cline{1-2}
A & \footnotesize \colorbox{blue!4.026}{\strut Mueller} \colorbox{blue!1.6019999999999999}{\strut agreed,} \colorbox{blue!3.994}{\strut and} \mybox{blue}{\colorbox{orange!5.133}{\strut Phillips}} \mybox{red}{\colorbox{orange!23.349}{\strut managed}} \mybox{green}{\colorbox{orange!2.309}{\strut Apollo}} \colorbox{orange!16.026}{\strut from} \colorbox{orange!8.782}{\strut January} \colorbox{orange!16.729}{\strut 1964,} \colorbox{orange!37.486999999999995}{\strut until} \colorbox{blue!1.7409999999999999}{\strut it} \colorbox{blue!8.558}{\strut achieved} \colorbox{blue!2.078}{\strut the} \colorbox{blue!6.77}{\strut first} \colorbox{blue!4.791}{\strut manned} \colorbox{blue!4.271}{\strut landing} \colorbox{blue!1.841}{\strut in} \colorbox{orange!5.708}{\strut July} \colorbox{orange!4.5409999999999995}{\strut 1969,} \colorbox{orange!3.95}{\strut after} \colorbox{orange!0.28600000000000003}{\strut which} \colorbox{orange!3.398}{\strut he} \colorbox{blue!2.905}{\strut returned} \colorbox{orange!1.719}{\strut to} \colorbox{orange!0.23700000000000002}{\strut Air} \colorbox{blue!2.5229999999999997}{\strut Force} \colorbox{blue!2.945}{\strut duty.} &  \\
\hline

\end{tabular}
\caption{
A RoBERTa-based model's correct prediction of \class{entailment} on the original input pair (Q, A) remains unchanged when the question is randomly shuffled (Q$_{1}$ \& Q$_{2}$) or when two random nouns in the question are swapped (Q$_{s}$).
The salient words in the questions \eg \colorbox{orange!24.573999999999998}{\strut manage} and \colorbox{blue!20.738}{\strut missions} remain similarly important after their context has been shuffled.
Also, the classifier harnessed self-attention to detect the correspondence between similar words that appear in both the question and the answer e.g. \mybox{red}{\colorbox{orange!27.526}{\strut manage}} (Q) and \mybox{red}{\colorbox{orange!21.132}{\strut managed}} (A).
That is, the top-3 pairs of words that were assigned the largest question-to-answer weights in a self-attention matrix (layer 0, head 7) are inside in the \textcolor{red}{red}, \textcolor{green}{green}, and \textcolor{blue}{blue} rectangles.
}
\label{fig:qnli_att_matching}
\end{figure*}

\subsubsection{Self-attention layers matching similar words in both the question and the answer}
\label{sec:qnli}

For sequence-pair tasks, \eg QNLI, how can models correctly predict \class{entailment} when the question words are randomly shuffled (Fig.~\ref{fig:qnli_att_matching}; Q$_1$) or when the question syntax is correct but its meaning changes entirely (Fig.~\ref{fig:qnli_att_matching}; Q$_s$).
We hypothesize that inside the model, there might be a self-attention (SA) layer that extracts pairs of similar words that appear in both the question and the answer (\eg ``manage'' vs. ``managed'' in Fig.~\ref{fig:qnli_att_matching}).

\subsec{Experiments} To test this hypothesis, we analyzed the 5,000 QNLI \devr examples (Table~\ref{table:preprocessing_stats}) of RoBERTa-based classifiers trained on QNLI.
For each example, we identified one SA matrix (among all 144 as the base model has 12 layers \& 12 heads per layer) that assigns the highest weights to pairs of similar words between the question and the answer, \ie excluding intra-question and intra-answer attention weights (see the procedure in Sec.~\ref{sec:qnli_appendix}).

\subsec{Results} First, in $\sim$58\% of the examples, we found at least three pairs of words that \emph{match} (\ie the sum Levenshtein character-level edit-distance for all 3 pairs is $\leq$ 4).
Second, we found, in total, 15 SA heads (out of the 144) which are explicitly tasked with capturing such question-to-answer word correspondence, \emph{regardless} of word order (see Fig.~\ref{fig:qnli_att_matching}).

Remarkably, \textbf{87\% of the work of matching similar words that appear in both the QNLI question and the answer was handled by only 3 self-attention heads at (layer, head) of (0,7), (1,9), and (2,6)}.

We found consistent results when repeating the same analysis for other three sequence-pair tasks.
That is, interestingly, \textbf{the three SA heads at exactly the same location of (0, 7), (1, 9), and (2, 6) account for 76\%, 89\%, and 83\% of the ``word-matching'' task on QQP, RTE, and MRPC, respectively}.
This coincidence is likely due to the fact that these classifiers were finetuned for different downstream tasks starting from the same pretrained RoBERTa encoder.
See Figs.~\ref{fig:teaser}, \ref{fig:qnli_att_matching}, \ref{fig:qualitative_examples_mrpc}--\ref{fig:qualitative_examples_rte} for qualitative examples of these three tasks.

\paragraph{How important are the 15 word-matching attention heads to QNLI model performance?}

We found that zero-ing out 15 random heads had almost no effect to correctly-classified predictions--\ie accuracy dropped marginally ($-1$\% to $-3$\%, Table \ref{table:zero_out_exp}) 
across different groups of examples.
However, ablating the 15 word-matching heads caused the performance to drop substantially 
\ie (a) by 9.6\% on the 1,453 \class{positive} examples identified in Sec.~\ref{sec:qnli_appendix}; (b) by 22.1\% on a set of 2,906 random, examples including both \class{positive} and \class{negative} examples (at 50/50 ratio); and (c) by 24.5\% on the entire QNLI 5,000-example \devr set.
That is, \textbf{the 15 SA heads that learned to detect similar words played an important role in solving QNLI, \ie enabling at least $\sim$50\% of the correct predictions} 
(Table~\ref{table:zero_out_exp}d; accuracy dropped from 100\% to 75.54\% when the random chance is 50\%).
In sum, we found overlap between words in the question and answer of QNLI examples and strong evidence that QNLI models harnessed self-attention to exploit such overlap to make correct decisions in spite of a random word-order.

\begin{table}[t]
\centering\small
\setlength\tabcolsep{0.6pt}
\begin{tabular}{lccc}
\toprule
QNLI \devr examples & \multirow{2}{*}{\makecell{Full \\ network}} & \multicolumn{2}{c}{Zero-out 15} \\
&  & \multicolumn{2}{c}{attention matrices} \\
\cmidrule(l{2pt}r{2pt}){3-4}
&  & Random & Ours \\
\toprule
a. 1,453 selected 0/5 (+) examples & 100 & 99.31 & 90.43 \\
\cmidrule{1-4}
b. 1,453 random 0/5 (+) examples & 100 & 99.24 & 91.05 \\
\cmidrule{1-4}
\makecell[l]{c. 1,453 random 0/5 (+) examples \\ ~~\& 1,453 random 0/5 (-) examples} & 100 & 98.18 & 77.91 \\
\cmidrule{1-4}
d. (+/-) All 5,000 examples & 100 & 96.96 & 75.54 \\
\bottomrule
\end{tabular}
\caption{
Zero-ing out a set of 15 ``word-matching'' self-attention matrices (identified via the procedure in Sec.~\ref{sec:qnli}) caused a substantial drop of $\sim$25\% in accuracy (d) while the random baseline is 50\%.
These 15 matrices played an important role in QNLI because ablating 15 random matrices only caused a $\sim$1-3\% drop in accuracy.
}
\label{table:zero_out_exp}
\end{table}

\subsection{Does increasing word-order sensitivity lead to higher model performance?}
\label{sec:extra_finetuning}

Here, we test whether encouraging BERT representations to be more sensitive to word order (\ie more syntax-aware) would improve model performance on GLUE \& SQuAD 2.0 \cite{rajpurkar2018know}.
We performed this test on the five GLUE binary-classification tasks (\ie excluding CoLA because its WOS score is already at 0.99; Table~\ref{table:miniQ1}).

\paragraph{Experiments}
Inspired by the fact that CoLA models are highly sensitive to word order, we finetuned the pretrained RoBERTa on a \emph{synthetic}, CoLA-like task first, before finetuning the model on downstream tasks.

The synthetic task is to classify a single sentence into \class{real} vs. \class{fake} where the latter is formed by taking each real sentence and swapping two random words in it.
For every downstream task (\eg SST-2), we directly used its original training and dev sets to construct a balanced, 2-class, synthetic dataset. 
After finetuning the pretrained RoBERTa on this synthetic binary classification task, we re-initialized the classification layer (keeping the rest unchanged) and continued finetuning it on a downstream task.

For both finetuning steps, we trained 5 models per task and followed the standard BERT finetuning procedure (described in Sec.~\ref{sec:classifiers}).

\paragraph{Results}

After the first finetuning on synthetic tasks, all models obtained a $\sim$99\% training-set accuracy and a $\sim$95\% dev-set accuracy.
\textbf{After the second finetuning on downstream tasks, we observed that all models were substantially more sensitive to word order}, compared to the baseline models (which were only finetuned on the downstream tasks).
That is, we repeated the 1-gram shuffling test (Sec.~\ref{sec:wos}) and found a $\sim$1.5 to $2\times$ increase in the WOS scores of all models (see Table \ref{table:new_miniQ1}a vs. b).

\begin{table}[h]
\centering\small
\setlength\tabcolsep{1.5pt}
\begin{tabular}{lcccc}
\toprule
GLUE \devs & \multicolumn{2}{c}{(a) RoBERTa} & \multicolumn{2}{c}{(b) Ours} \\
\cmidrule(l{3pt}r{3pt}){2-3} 
\cmidrule(l{3pt}r{3pt}){4-5}
& Accuracy & WOS & Accuracy & WOS  \\
\toprule 
\texttt{RTE} & 80.76 & 0.38 & 64.01 & \textbf{0.72} (+189\%) \\
\cmidrule{1-5}
\texttt{MRPC} & 83.86 & 0.32 & 72.88 & \textbf{0.54} (+169\%) \\
\cmidrule{1-5}
\texttt{SST-2} & 84.26 & 0.31 & 76.97 & \textbf{0.46} (+148\%) \\
\cmidrule{1-5}
\texttt{QQP} & 87.66 & 0.25 & 77.11 & \textbf{0.46} (+184\%) \\
\cmidrule{1-5}
\texttt{QNLI} & 91.09 & 0.18 & 82.44 & \textbf{0.35} (+194\%) \\
\bottomrule
\end{tabular}
\caption{
With finetuning on synthetic tasks, all of our models (b) have a larger drop in accuracy 
on shuffled \devs examples, compared to the standard RoBERTa-based classifiers (a).
That is, our models are substantially more sensitive to word-order randomization (\ie +148\% to +194\% in WOS scores).
}
\label{table:new_miniQ1}
\end{table}


\paragraph{GLUE} On GLUE dev sets, on average over 5 runs, our models outperformed the RoBERTa baseline on all tasks except for SST-2 (Table~\ref{table:extra_finetuning}).
The highest improvement is in RTE (from 72.2\% to 73.21\% on average, and to 74.73\% for the best single model), which is consistent with the fact that RTE has the highest WOS score among non-CoLA tasks (Sec.~\ref{sec:wos}).

\paragraph{SQuAD 2.0}
Our models also outperformed the RoBERTa baseline on the SQuAD 2.0 dev set, with the highest F1 gain from 80.62\% to 81.08\% (Table~\ref{table:extra_finetuning}).

In sum, leveraging the insights that the original BERT-based models are largely word-order invariant, we showed that increasing model sensitivity via a simple extra finetuning step directly improves GLUE and SQuAD 2.0 performance.

\begin{table}[ht]
\centering\small
\setlength\tabcolsep{3.2pt}
\begin{tabular}{lcccccc}
\toprule 
& \multirow{2}{*}{RTE} & \multirow{2}{*}{QQP} & \multirow{2}{*}{MRPC} & \multirow{2}{*}{SST-2} & \multirow{2}{*}{QNLI} & \multirow{2}{*}{SQuAD} \\
&  &  &  &  &  &  \\
& (Acc) & (Acc) & (Acc) & (Acc) & (Acc) & (F1) \\
\toprule
RoBERTa & 72.20 & 91.12 & 87.25 & \textbf{94.50} & 92.57 & 80.62 \\
\cmidrule{0-6}
Our best & \textbf{74.73} & \textbf{91.31} & \textbf{88.73} & 94.50 & \textbf{93.08} & \textbf{81.08} \\
model & \textbf{+2.53} & \textbf{+0.19} & \textbf{+1.48} & +0 & \textbf{+0.51} & \textbf{+0.46} \\
\cmidrule{0-6}
Average & \textbf{73.21} & \textbf{91.19} & \textbf{87.31} & 94.22 & \textbf{92.71} & \textbf{80.75} \\
(5 runs) & \textbf{+1.01} & \textbf{+0.07} & \textbf{+0.06} & -0.28 & \textbf{+0.14} & \textbf{+0.13} \\
\bottomrule
\end{tabular}
\caption{
Finetuning the pretrained RoBERTa on synthetic tasks (before finetuning on the downstream tasks) improved model dev-set performance on SQuAD 2.0 (b) and all the tested tasks in GLUE (a), except SST-2.
}
\label{table:extra_finetuning}
\end{table}

\section{Related Work}

\subsec{Pretrained BERT} 
\citet{lin2019open} found that positional information is encoded in the first there layers of \bertBase and fades out starting layer 4.
\citet{ettinger2020bert} found that BERT heavily relies on word order when predicting missing words in masked sentences from the CPRAG-102 dataset. 
That is, shuffling words in the context sentence caused the word-prediction accuracy to drop by $\sim$1.3 to 2$\times$.
While all above work studied the \emph{pretrained} BERT, we instead study BERT-based models \emph{finetuned} on downstream tasks.

\paragraph{Word-ordering as an objective}
In text generation, \citet{elman1990finding} found that recurrent neural networks were sensitive to regularities in word order in simple sentences.
Language models~\citep{mikolov2010recurrent} with long short-term memory (LSTM) units \citep{hochreiter1997long} were able to recover the original word order of a sentence from randomly-shuffled words even without any explicit syntactic information \citep{schmaltz2016word}.
\citealt{wang2020structbert} also observed an increase in GLUE performance after pretraining BERT with two additional objectives of word-ordering and sentence-ordering.
Their work differs from ours in three points: (1) they did not study the importance of word order alone; (2) StructBERT improvements were inconsistent across tasks and models (Table~\ref{table:miniQ1}d) and motivated us to compare the word-order importance between GLUE tasks; and (3) we proposed to improve model performance by finetuning not pretraining.


\paragraph{Word-order insensitivity in other NLP tasks}
ML models have been shown to be insensitive to word order in several NLP tasks such as reading comprehension \cite{si2019does,sugawara2020assessing}, dialog \cite{sankar-etal-2019-neural}, natural language inference \cite{parikh-etal-2016-decomposable,sinha2020unnatural}, and essay scoring \cite{parekh2020my}.
\citealt{zanzotto2020kermit} found that for several text classification tasks, syntactic information was not always required.
In word prediction, LSTMs and pre-trained BERT were found to exhibit a certain degree of insensitivity when the context words are randomly shuffled \cite{khandelwal-etal-2018-sharp,mitchell2020priorless,ettinger2020bert}.
Compared to the prior work, we are the first to perform a word-order analysis on a NLU benchmark and to contrast this sensitivity across the tasks.


\paragraph{Humans can also be word-order invariant}
A recent human study interestingly showed that sentences with scrambled word orders elicit a response as high as that elicited by original sentences as long as the local mutual information among words is high enough \cite{mollica2020composition}.
\citealt{gibson2013rational} found that humans can also exhibit word-order-invariance effects, especially when one interpretation is much more semantically plausible.
Our work therefore documents an important similarity between humans and advanced NLU models.



\paragraph{Invariance to patch-order in computer vision}
In computer vision, the accuracy of state-of-the-art image classifiers was found to only drop marginally when the patches in an image were randomly shuffled \cite{chen2020shape,zhang2019interpreting}.

\section{Discussion and Conclusion}

Consistently across three BERT variants and two model sizes, we found that GLUE-trained BERT-based models are often word-order invariant unless explicitly asked for (\eg in CoLA).

We present a reflection on the progress of NLU by studying GLUE---a benchmark where humans have been surpassed by many models in the past 18 months.
As suggested by our work, these models; however, may neither use syntactic information nor complex reasoning.
We revealed how self-attention, a key building block in modern NLP, is being used to extract superficial cues to solve sequence-pair GLUE tasks even when words are out of order.

\paragraph{Adversarial NLI} 
We also replicated our shuffling experiments on ANLI \cite{nie-etal-2020-adversarial}, a task considered challenging to existing models, and where RoBERTa-based models only obtained a 56\% accuracy.
We found RoBERTa-based models to remain not always sensitive to word-order randomization on ANLI (Table \ref{table:miniQ1_anli}; WOS of 0.63), suggesting a common issue in existing benchmarks.

\subsubsection*{Acknowledgments}
We thank Michael Alcorn, Qi Li, and Peijie Chen for helpful feedback on the early results. 
We are grateful for valuable feedback from Sam Bowman, Ernest Davis, and Melanie Mitchell.
AN is supported by the National Science Foundation under Grant No. 1850117, and donations from the NaphCare Charitable Foundation, and Nvidia.

\clearpage
\newpage
\bibliography{references}
\bibliographystyle{_ACL2021/acl_natbib}

\appendix



\newcommand{\beginsupplementary}{
    \setcounter{table}{0}
    \renewcommand{\thetable}{A\arabic{table}}
    \setcounter{figure}{0}
    \renewcommand{\thefigure}{A\arabic{figure}}		\setcounter{section}{0}
}
\beginsupplementary




\clearpage
\newpage

\section{Self-attention layers that match question-words to similar words in the answer}
\label{sec:qnli_appendix}

QNLI models being so insensitive to word shuffling (\ie 89.4\% of the correct predictions remain correct) suggests that inside the finetuned BERT, there might be a self-attention (SA) layer that extract pairs of similar words that appear in both the question and answer.

We started by analyzing all 2,500 \class{positive} \devr examples (Table~\ref{table:preprocessing_stats}) of RoBERTa-based classifiers trained on QNLI because there were fewer and more consistent ways for labeling a sentence \class{positive} than for the \class{negative} (shown in Sec.~\ref{sec:heatmap_similarity}).

\paragraph{Experiment}
There were 1,776 (out of 2,500) examples whose predictions did not change in 5 random shufflings (a.k.a 5/5 subset).
For each such example, we followed the following 4 steps to identify one SA matrix (among all 144 as the base model has 12 layers \& 12 heads per layer) that captures the strongest attention connecting the question and answer words.

\begin{enumerate}
    \item Per example $x$, we created its shuffled version $\hat{x}$ by randomly shuffling words in the question and fed $\hat{x}$ into the classifier.
    \item For each SA matrix obtained, we identified the top-3 highest-attention weights that connect the shuffled question tokens and the real answer tokens (\ie excluding attention weights between question tokens or answer tokens only).
    \item For each shuffled example $\hat{x}$, we identified one matrix $M$ whose the top-3 word pairs are the nearest in Levenshtein character-level edit-distance \citep{nltk2020edit}.
    For instance, the distance is 1 between \mybox{red}{\colorbox{orange!27.526}{\strut manage}} and \mybox{red}{\colorbox{orange!21.132}{\strut managed}} (Fig.~\ref{fig:qnli_att_matching}).
    \item For each matrix $M$ identified for $\hat{x}$, we fed the corresponding real example $x$ through the network and re-computed the edit-distance for each of the top-3 word pairs.
\end{enumerate}

\begin{figure*}[t]
	\centering
	\begin{subfigure}{0.65\linewidth}
		\centering
		\includegraphics[width=1.0\linewidth]{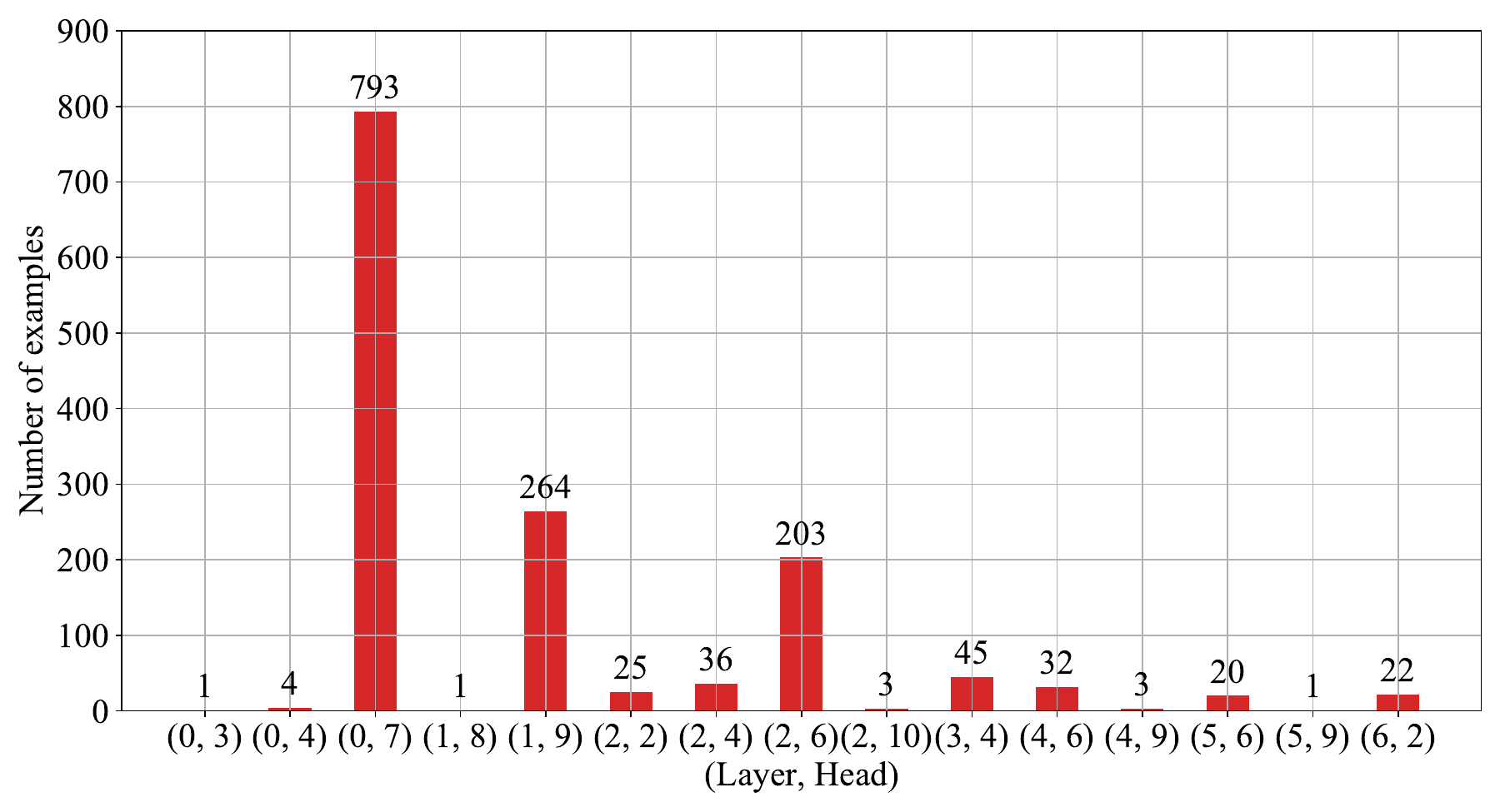}
		\caption{Histogram of self-attention matrices}
		\label{fig:self_attention_histogram}
	\end{subfigure}
	\hspace{1mm}
	\begin{subfigure}{0.33\linewidth}
		\centering
		\small
        \setlength\tabcolsep{1pt}
        \resizebox{\textwidth}{!}{
        \begin{tabular}{lrr}
        \toprule
        $($Layer, Head$)$ & \# Examples & \% \\
        \toprule 
        \makecell[l]{$($0, 3$)$, $($0, 4$)$, $($0, 7$)$} & 798 & 54.9\% \\ \midrule
        $($1, 8$)$, $($1, 9$)$ & 265 & 18.2\% \\ \midrule
        \makecell[l]{$($2, 2$)$, $($2, 4$)$, $($2, 6$)$, $($2, 10$)$} & 267 & 18.4\% \\ \midrule
        $($3, 4$)$ & 45 & 3.1\% \\ \midrule
        $($4, 6$)$, $($4, 9$)$ & 35 & 2.4\% \\ \midrule
        $($5, 6$)$, $($5, 9$)$ & 21 & 1.5\% \\ \midrule
        $($6, 2$)$ & 22 & 1.5\% \\ 
        \midrule
        Total & 1,453 & 100\% \\
        \bottomrule
        \end{tabular}
        }
		\caption{Layer-wise comparison}
		\label{fig:self_attention_table}
	\end{subfigure}
	\caption{
	    Among 144 self-attention matrices in the RoBERTa-based classifier finetuned for QNLI, there are 15 ``word-matching'' matrices (a) that explicitly attend to pairs of similar words that appear in both questions and answers regardless of the order of words in the question (see example pairs in Fig.~\ref{fig:qnli_att_matching}).
	    For each QNLI example, we identified one such matrix that exhibits the matching behavior the strongest (a).
	    92\% of the task of attending to duplicate words is mostly handled in the first three layers (b).
	}
	\label{fig:qnli_att_layer_head_stats}
\end{figure*}

\paragraph{Results}
At step 3, there were 1,590 SA matrices (out of 1,776) whose the top-3 SA weights connected three pairs of \emph{matching} words (\ie the total edit-distance for 3 pairs together is $\leq$ 4)\footnote{4 is a tight budget to account for minor typos or punctuation differences \eg ``Amazon'' vs. ``Amazon's''.} that appear in both the shuffled question and original answer (see example top-3 pairs in Fig.~\ref{fig:qnli_att_matching}).
At step 4, this number only dropped slightly to 1,453 matrices when replacing the shuffled question by the original one (see Table~\ref{table:qnli_sumedit_distance} for detailed statistics).

\begin{table}[h]
\centering\small
\setlength\tabcolsep{5pt}
\begin{tabular}{crrrr}
\toprule
\multirow{2}{*}{\makecell{Sum\\distance}} & \multicolumn{2}{l}{(a) \devs alone} & \multicolumn{2}{c}{(b) \devs \& \devr} \\
\cmidrule(l{2pt}r{2pt}){2-3} \cmidrule(l{2pt}r{2pt}){4-5}
& \# examples & \% & \# examples & \% \\
\toprule

$\leq 0$ & 749 & 42.17 & 392 & 24.65 \\ \cmidrule(l{2pt}r{2pt}){1-5}
$\leq 1$ & 1,253 & 70.55 & 1,071 & 67.36 \\ \cmidrule(l{2pt}r{2pt}){1-5}
$\leq 2$ & 1,440 & 81.08 & 1,283 & 80.69 \\ \cmidrule(l{2pt}r{2pt}){1-5}
$\leq 3$ & 1,543 & 86.88 & 1,391 & 87.48 \\ \cmidrule(l{2pt}r{2pt}){1-5}
$\leq 4$ & \textbf{1,590} & \textbf{89.53} & \textbf{1,453} & \textbf{91.38} \\ \cmidrule(l{2pt}r{2pt}){1-5}
$\leq 15$ & 1,776 & 100.00 & 1,574 & 98.99 \\

\toprule
Total & 1,776 & 100.00 & 1,590 & 100.00 \\ 
\bottomrule
\end{tabular} 
\caption{
The number of QNLI examples where we found $\geq$ one self-attention matrix that the most strongly attends to three pairs of \emph{matching} words when given the \devs examples \ie (modified question, real answer) (a) or when given both the shuffled and real examples (b).
In other words, \textbf{the numbers in (b) denote the number of examples where (1) there exist $\geq$ 3 words, regardless of its word order, in the question that can be found in the accompanying real answer; and (2) these correspondences are captured by at least one self-attention matrix.}
The sum edit-distance for all 3 pairs of words are less than $N$ where $N = \{0, 1, 2, 3, 4, 15\}$ (left column).
}
\label{table:qnli_sumedit_distance}
\end{table}

However, there are only 15 unique, RoBERTa self-attention matrices in these 1,453 examples (see Fig.~\ref{fig:qnli_att_layer_head_stats}).
Also at step 4, 83\% of the same word pairs remained within the top-3 of the same SA matrices, after question replacement, \ie 17\% of attention changed to different pairs \eg from (``managed'', ``manage'') to (``it'', ``it'').

First, our results showed that there is a set of 15 self-attention heads explicitly tasked with capturing question-to-answer word correspondence \emph{regardless of word order}.
Second, \textbf{for $\sim$58\% (\ie 1,453 / 2,500) of QNLI \class{positive} examples: (1) there exist $\geq$ 3 words in the question that can be found in the accompanying answer; and (2) these correspondences are captured by at least one of the 15 SA matrices.}
We also found similar results for 2,500 \class{negative} \devr examples (data not shown).

\begin{table*}[t]
\centering\small
\setlength\tabcolsep{3pt}
\begin{tabular}{llcccccccccccc}
\toprule
Model & Task & \multicolumn{2}{c}{(a) Perf. on \devr} & \multicolumn{4}{c}{(b) Perf. on \devs} & \multicolumn{4}{c}{(c) Word-Order Sensitivity} \\
\cmidrule(l{1pt}r{1pt}){3-4} \cmidrule(l{1pt}r{1pt}){5-8} \cmidrule(l{1pt}r{1pt}){9-12}
&  & Models & Baseline & 2-noun swap & 1-gram & 2-gram & 3-gram & 2-noun swap & 1-gram & 2-gram & 3-gram\\
\toprule 
\robertaBase & \texttt{ANLI}  & 100 & 33.33 & 74.26 & 57.74 & 66.63 & 69.04 & 0.39 & 0.63 & 0.50 & 0.46 \\
\cmidrule{2-12}
& \texttt{A1}  & 100 & 33.33 & 81.46 & 63.31 & 71.52 & 75.37 & 0.28 & 0.55 & 0.43 & 0.37 \\
\cmidrule{2-12}
& \texttt{A2}  & 100 & 33.33 & 70.83 & 54.61 & 64.73 & 67.02 & 0.44 & 0.68 & 0.53 & 0.49 \\
\cmidrule{2-12}
& \texttt{A3}  & 100 & 33.33 & 70.50 & 55.29 & 63.63 & 64.73 & 0.55 & 0.67 & 0.55 & 0.53 \\
\cmidrule{1-12}
\robertaLarge & \texttt{ANLI}  & 100 & 33.33 & 70.41 & 54.87 & 64.11 & 68.76 & 0.44 & 0.68 & 0.54 & 0.47 \\
\cmidrule{2-12}
& \texttt{A1}  & 100 & 33.33 & 78.06 & 60.31 & 70.57 & 75.86 & 0.33 & 0.6 & 0.44 & 0.36 \\
\cmidrule{2-12}
& \texttt{A2}  & 100 & 33.33 & 67.88 & 51.44 & 60.64 & 66.31 & 0.48 & 0.73 & 0.59 & 0.51 \\
\cmidrule{2-12}
& \texttt{A3}  & 100 & 33.33 & 65.30 & 52.85 & 61.11 & 64.10 & 0.52 & 0.71 & 0.58 & 0.54 \\
\bottomrule
\end{tabular}
\caption{
All results (a--c) of \robertaBase and \robertaLarge models finetuned on the combination of NLI datasets (SNLI, MNLI, FEVER and ANLI) are reported on the ANLI \devr sets (\ie 100\% accuracy) which includes A1, A2 and A3 (a).
The accuracies for \robertaBase and \robertaLarge on ANLI are 51.19\% and 56.98\%, respectively.
Each row is computed by averaging the results of 10 random shuffles.
Word-Order Sensitivity (WOS) of ANLI and its subsets (c).
Since ANLI is 3-way classification task, the baseline is 33.33\% (as described in Sec~\ref{sec:dataset_construction}).
}
\label{table:miniQ1_anli}
\end{table*}

\begin{table*}[hbt!]
\centering\small
\setlength\tabcolsep{3pt}
\begin{tabular}{ccccc}
\toprule
\multicolumn{1}{c}{Dictionary} & \multicolumn{2}{c}{Opinion Lexicon \cite{hu2004mining}} & \multicolumn{2}{c}{SentiWords \cite{gatti2015sentiwords}} \\
\cmidrule(l{2pt}r{2pt}){2-3} \cmidrule(l{2pt}r{2pt}){4-5}
& \multicolumn{1}{c}{(a) RoBERTa} & \multicolumn{1}{c}{(b) ALBERT} & \multicolumn{1}{c}{(c) RoBERTa} & \multicolumn{1}{c}{(d) ALBERT} \\
\toprule

Total examples in subset 5/5 & \multicolumn{1}{c}{523} & \multicolumn{1}{c}{506} & \multicolumn{1}{c}{523} & \multicolumn{1}{c}{506} \\
(\class{positive} / \class{negative}) & \multicolumn{1}{c}{(278 / 245)} & \multicolumn{1}{c}{(228 / 278)} & \multicolumn{1}{c}{(278 / 245)} & \multicolumn{1}{c}{(228 / 278)} \\

\cmidrule(l{2pt}r{2pt}){1-5}
Not found in dictionary & \makecell{223 / 523 \\ (42.64\%)} & \makecell{217 / 506 \\ (42.89\%)} & \makecell{110 / 523 \\ (21.03\%)} & \makecell{104 / 506 \\ (20.55\%)} \\
\cmidrule(l{2pt}r{2pt}){1-5}
Found in dictionary & \makecell{300 / 523 \\ (57.36\%)} & \makecell{289 / 506 \\ (57.11\%)} & \makecell{413 / 523 \\ (78.97\%)} & \makecell{402 / 506 \\ (79.45\%)} \\
\cmidrule(l{2pt}r{2pt}){1-5}
P ( \class{positive} sentence $\mid$ positive top-1 word )  & \makecell{174 / 174 \\ (100.00\%)} & \makecell{143 / 144 \\ (99.31\%)} & \makecell{222 / 258 \\ (86.05\%)} & \makecell{186 / 215 \\ (86.51\%)} \\
\cmidrule(l{2pt}r{2pt}){1-5}
P ( \class{negative} sentence $\mid$ negative top-1 word ) & \makecell{119 / 126 \\ (94.44\%)} & \makecell{136 / 145 \\ (93.79\%)} & \makecell{145 / 155 \\ (93.55\%)} & \makecell{177 / 187 \\ (94.65\%)} \\

\bottomrule
\end{tabular}
\caption{
\textbf{If the top-1 most important word in an SST-2 5/5 example has a positive meaning, then there is a 100\% chance that the sentence is labeled \class{positive} in SST-2.}
Similarly, the conditional probability of a sentence being labeled \class{negative} given a negative most important word (by LIME \citealt{ribeiro2016should}) is 94.44\%.
}
\label{table:sst2_lexicons}
\end{table*}

\clearpage
\newpage

\begin{figure*}[ht]
\centering\small 
\setlength\tabcolsep{2pt} 

\begin{tabular}{|l|l|c|}
\hline
\multicolumn{2}{|c}{\multirow{2}{*}{LIME attributions (\colorbox{blue!100}{\strut \textcolor{white}{negative -1}}, neutral 0, \colorbox{orange!100}{\strut positive +1})} } & \multirow{2}{*}{\makecell{ }} \\
\multicolumn{2}{|c}{\multirow{2}{*}{}} & \\
\hline
\multicolumn{3}{|l|}{\cellcolor{white} \textbf{CoLA} example.~ Groundtruth: \class{acceptable}} \\
\hline

S & \footnotesize \colorbox{orange!12.777}{\strut Medea} \colorbox{blue!0.757}{\strut denied} \colorbox{orange!14.018}{\strut poisoning} \colorbox{orange!5.56}{\strut the} \colorbox{blue!4.989}{\strut phoenix.} & \class{acceptable} 0.99 \\
\hline
\hline
S$_{1}$ & \footnotesize \colorbox{blue!10.389}{\strut poisoning} \colorbox{blue!11.198}{\strut the} \colorbox{orange!11.047}{\strut phoenix} \colorbox{blue!13.618}{\strut denied} \colorbox{blue!7.7090000000000005}{\strut Medea.} & \class{acceptable} 0.53 \\
\hline
S$_{2}$ & \footnotesize \colorbox{orange!1.6219999999999999}{\strut phoenix} \colorbox{orange!5.225}{\strut Medea} \colorbox{orange!16.377}{\strut denied} \colorbox{orange!8.488999999999999}{\strut the} \colorbox{orange!3.2710000000000004}{\strut poisoning.} & \class{acceptable} 0.99 \\
\hline
S$_{3}$ & \footnotesize \colorbox{orange!11.094999999999999}{\strut Medea} \colorbox{blue!9.127}{\strut the} \colorbox{orange!6.920999999999999}{\strut poisoning} \colorbox{blue!5.331}{\strut phoenix} \colorbox{blue!37.464999999999996}{\strut denied.} & \class{unacceptable} 0.95 \\
\hline
S$_{4}$ & \footnotesize \colorbox{orange!2.809}{\strut phoenix} \colorbox{orange!3.8}{\strut Medea} \colorbox{orange!16.843}{\strut denied} \colorbox{orange!8.846}{\strut the} \colorbox{orange!0.9079999999999999}{\strut poisoning.} & \class{unacceptable} 0.99 \\
\hline
S$_{5}$ & \footnotesize \colorbox{orange!1.957}{\strut Medea} \colorbox{blue!7.949000000000001}{\strut phoenix} \colorbox{orange!10.676}{\strut poisoning} \colorbox{blue!11.91}{\strut the} \colorbox{blue!16.896}{\strut denied.} & \class{unacceptable} 0.96 \\
\hline
\end{tabular}

\caption{Each CoLA example contains a single sentence.
Here, we shuffled the words in the original sentence (S) five times to create five new sentences (S$_{1}$ to S$_{5}$) and fed them to a RoBERTa-based classifier for predictions.
Words that are important for or against the prediction (by LIME \citealt{ribeiro2016should}) are in \colorbox{orange!44.756}{\strut orange} and \colorbox{blue!44.756}{\strut blue}, respectively.
Most of the shuffled examples were classified into \class{unacceptable} label (\ie grammatically incorrect) with even higher confidence score than the original ones.
}
\label{fig:qualitative_examples_cola}

\end{figure*}


\begin{figure*}[ht]
\centering\small 
\setlength\tabcolsep{2pt}

\begin{tabular}{|l|p{0.77\linewidth}|c|}
\hline

\multicolumn{3}{|l|}{\cellcolor{white} \textbf{MRPC} example.~ Groundtruth: \class{equivalent}} \\
\hline

A & \footnotesize \colorbox{orange!5.815}{\strut My} \mybox{blue}{\colorbox{orange!6.3229999999999995}{\strut decision}} \colorbox{orange!1.0330000000000001}{\strut today} \colorbox{orange!2.331}{\strut is} \colorbox{orange!9.209}{\strut not} \mybox{green}{\colorbox{orange!12.352}{\strut based}} \colorbox{orange!5.899}{\strut on} \colorbox{orange!14.112}{\strut any} \colorbox{orange!11.713}{\strut one} \mybox{red}{\colorbox{orange!14.493}{\strut event}} \colorbox{orange!0}{\strut .} \colorbox{orange!0}{\strut "} & \multirow{2}{*}{\class{equivalent} 0.99} \\
\cline{1-2}
B & \footnotesize \colorbox{blue!1.725}{\strut Governor} \colorbox{blue!1.149}{\strut Rowland} \colorbox{orange!2.012}{\strut said} \colorbox{orange!2.4410000000000003}{\strut his} \mybox{blue}{\colorbox{orange!9.372}{\strut decision}} \colorbox{orange!3.26}{\strut was} \colorbox{orange!0}{\strut "} \colorbox{orange!6.318}{\strut not} \mybox{green}{\colorbox{orange!9.364}{\strut based}} \colorbox{orange!7.005}{\strut on} \colorbox{orange!13.972000000000001}{\strut any} \colorbox{orange!11.162999999999998}{\strut one} \mybox{red}{\colorbox{orange!10.0}{\strut event}} \colorbox{orange!0}{\strut .} \colorbox{orange!0}{\strut "} &  \\
\hline\hline
A$_1$ & \footnotesize \text{\mybox{green}{\colorbox{orange!16.477}{\strut event}}} \colorbox{orange!13.577}{\strut any} \colorbox{orange!0.254}{\strut is} \colorbox{orange!21.353}{\strut one} \mybox{red}{\colorbox{orange!14.030999999999999}{\strut decision}} \mybox{blue}{\colorbox{orange!28.898000000000003}{\strut based}} \colorbox{orange!10.448}{\strut on} \colorbox{orange!2.458}{\strut My} \colorbox{orange!2.793}{\strut today} \colorbox{orange!15.516}{\strut not} \colorbox{orange!0}{\strut .} \colorbox{orange!0}{\strut "} & \multirow{2}{*}{\class{equivalent} 0.98} \\
\cline{1-2}
B & \footnotesize \colorbox{blue!4.642}{\strut Governor} \colorbox{blue!2.995}{\strut Rowland} \colorbox{orange!3.614}{\strut said} \colorbox{orange!4.836}{\strut his} \mybox{red}{\colorbox{orange!19.628}{\strut decision}} \colorbox{orange!5.257}{\strut was} \colorbox{orange!0}{\strut "} \colorbox{orange!9.524000000000001}{\strut not} \mybox{blue}{\colorbox{orange!18.864}{\strut based}} \colorbox{orange!9.951}{\strut on} \colorbox{orange!17.019000000000002}{\strut any} \colorbox{orange!15.22}{\strut one} \mybox{green}{\colorbox{orange!1.891}{\strut event}} \colorbox{orange!0}{\strut .} \colorbox{orange!0}{\strut "} &  \\
\hline
\hline
A$_{2}$ & \footnotesize \colorbox{orange!11.125}{\strut one} \mybox{red}{\colorbox{orange!19.064}{\strut based}} \colorbox{blue!3.778}{\strut today} \colorbox{orange!19.502}{\strut not} \colorbox{orange!23.515}{\strut any} \colorbox{orange!0.776}{\strut My} \colorbox{orange!8.988999999999999}{\strut on} \mybox{green}{\colorbox{orange!21.844}{\strut event}} \colorbox{orange!3.675}{\strut is} \mybox{blue}{\colorbox{orange!13.841000000000001}{\strut decision}} \colorbox{orange!0}{\strut .} \colorbox{orange!0}{\strut "} & \multirow{2}{*}{\class{equivalent} 0.98} \\
\cline{1-2}
B & \footnotesize \colorbox{blue!1.6260000000000001}{\strut Governor} \colorbox{blue!4.546}{\strut Rowland} \colorbox{orange!7.469}{\strut said} \colorbox{orange!1.425}{\strut his} \mybox{blue}{\colorbox{orange!26.471}{\strut decision}} \colorbox{orange!4.882000000000001}{\strut was} \colorbox{orange!0}{\strut "} \colorbox{orange!12.334}{\strut not} \mybox{red}{\colorbox{orange!20.346}{\strut based}} \colorbox{orange!8.593}{\strut on} \colorbox{orange!16.431}{\strut any} \colorbox{orange!3.433}{\strut one} \mybox{green}{\colorbox{orange!20.793}{\strut event}} \colorbox{orange!0}{\strut .} \colorbox{orange!0}{\strut "} &  \\
\hline
\end{tabular}

\caption{
Each MRPC example contains a pair of sentences \ie (A, B).
Here, we shuffled the words in the original sentence (A) to create modified sentences (A$_{1}$ \& A$_{2}$) and fed them together with the original second sentence (B) to a RoBERTa-based classifier for predictions.
Also, the classifier harnessed self-attention to detect the correspondence between similar words that appear in both sentences.
That is, the top-3 pairs of words that were assigned the largest cross-sentence weights in a self-attention matrix (layer 0, head 7) are inside in the \textcolor{red}{red}, \textcolor{green}{green}, and \textcolor{blue}{blue} rectangles.
}
\label{fig:qualitative_examples_mrpc}
\end{figure*}

\begin{figure*}[ht]
\centering\small 
\setlength\tabcolsep{2pt}

\begin{tabular}{|l|p{0.77\linewidth}|c|}
\hline

\multicolumn{3}{|l|}{\cellcolor{white}  \textbf{RTE} example.~ Groundtruth: \class{entailment}} \\
\hline

P & \footnotesize \colorbox{blue!2.948}{\strut About} \mybox{red}{\colorbox{orange!0}{\strut 33.5}} \colorbox{orange!2.734}{\strut million} \mybox{green}{\colorbox{orange!7.332}{\strut people}} \colorbox{orange!12.208}{\strut live} \colorbox{orange!3.8249999999999997}{\strut in} \colorbox{orange!4.289}{\strut this} \colorbox{orange!5.686999999999999}{\strut massive} \colorbox{orange!1.059}{\strut conurbation.} \colorbox{orange!1.095}{\strut I} \colorbox{blue!2.581}{\strut would} \colorbox{orange!0.849}{\strut guess} \colorbox{orange!3.614}{\strut that} \colorbox{blue!1.042}{\strut 95\%} \colorbox{orange!3.398}{\strut of} \colorbox{blue!1.11}{\strut the} \colorbox{orange!0}{\strut 5,000} \colorbox{orange!0.932}{\strut officially} \colorbox{orange!0}{\strut foreign-capital} \colorbox{blue!7.069}{\strut firms} \colorbox{orange!3.8249999999999997}{\strut in} \colorbox{orange!12.727}{\strut Japan} \colorbox{blue!1.966}{\strut are} \colorbox{orange!4.373}{\strut based} \colorbox{orange!3.8249999999999997}{\strut in} \mybox{blue}{\colorbox{orange!18.235}{\strut Tokyo.}} & \multirow{2}{*}{\class{entailment} 0.90} \\
\cline{1-2}
H & \footnotesize \colorbox{orange!4.587}{\strut About} \mybox{red}{\colorbox{orange!0}{\strut 33.5}} \colorbox{orange!5.095000000000001}{\strut miilion} \mybox{green}{\colorbox{orange!11.385000000000002}{\strut people}} \colorbox{orange!12.212}{\strut live} \colorbox{orange!4.445}{\strut in} \mybox{blue}{\colorbox{orange!11.597}{\strut Tokyo.}} &  \\
\hline
\hline
P & \footnotesize \colorbox{orange!9.658999999999999}{\strut About} \mybox{red}{\colorbox{orange!0}{\strut 33.5}} \colorbox{orange!7.172000000000001}{\strut million} \mybox{green}{\colorbox{orange!5.090999999999999}{\strut people}} \mybox{blue}{\colorbox{orange!15.98}{\strut live}} \colorbox{orange!7.9079999999999995}{\strut in} \colorbox{orange!1.9980000000000002}{\strut this} \colorbox{orange!8.247}{\strut massive} \colorbox{orange!3.843}{\strut conurbation.} \colorbox{orange!1.783}{\strut I} \colorbox{blue!2.497}{\strut would} \colorbox{orange!2.789}{\strut guess} \colorbox{orange!0.876}{\strut that} \colorbox{orange!0.396}{\strut 95\%} \colorbox{orange!0.244}{\strut of} \colorbox{orange!0.922}{\strut the} \colorbox{orange!0}{\strut 5,000} \colorbox{orange!1.6320000000000001}{\strut officially} \colorbox{orange!0}{\strut foreign-capital} \colorbox{blue!3.941}{\strut firms} \colorbox{orange!7.9079999999999995}{\strut in} \colorbox{orange!2.021}{\strut Japan} \colorbox{orange!0.633}{\strut are} \colorbox{orange!1.659}{\strut based} \colorbox{orange!7.9079999999999995}{\strut in} \colorbox{orange!16.029}{\strut Tokyo.} & \multirow{2}{*}{\class{entailment} 0.79} \\
\cline{1-2}
H$_{1}$ & \footnotesize \text{\mybox{green}{\colorbox{orange!5.1499999999999995}{\strut people}}} \colorbox{orange!2.3}{\strut in} \colorbox{orange!5.085}{\strut miilion} \mybox{red}{\colorbox{orange!0}{\strut 33.5}} \mybox{blue}{\colorbox{orange!12.956000000000001}{\strut live}} \colorbox{orange!12.159}{\strut Tokyo} \colorbox{orange!0}{\strut About.} &  \\
\hline\hline
P & \footnotesize \colorbox{orange!6.931}{\strut About} \mybox{red}{\colorbox{orange!0}{\strut 33.5}} \colorbox{orange!8.493}{\strut million} \mybox{green}{\colorbox{orange!7.71}{\strut people}} \mybox{blue}{\colorbox{orange!20.017}{\strut live}} \colorbox{orange!3.611}{\strut in} \colorbox{orange!5.196}{\strut this} \colorbox{orange!10.593}{\strut massive} \colorbox{orange!3.864}{\strut conurbation.} \colorbox{blue!0.055999999999999994}{\strut I} \colorbox{blue!0.28500000000000003}{\strut would} \colorbox{orange!1.419}{\strut guess} \colorbox{orange!1.0410000000000001}{\strut that} \colorbox{orange!1.8239999999999998}{\strut 95\%} \colorbox{orange!1.567}{\strut of} \colorbox{orange!0.976}{\strut the} \colorbox{orange!0}{\strut 5,000} \colorbox{orange!0.992}{\strut officially} \colorbox{orange!0}{\strut foreign-capital} \colorbox{blue!3.791}{\strut firms} \colorbox{orange!3.611}{\strut in} \colorbox{orange!4.7010000000000005}{\strut Japan} \colorbox{blue!0.154}{\strut are} \colorbox{orange!3.4619999999999997}{\strut based} \colorbox{orange!3.611}{\strut in} \colorbox{orange!13.494}{\strut Tokyo.} & \multirow{2}{*}{\class{entailment} 0.80} \\
\cline{1-2}
H$_{2}$ & \footnotesize \text{\mybox{red}{\colorbox{orange!0}{\strut 33.5}}} \colorbox{blue!0.932}{\strut in} \mybox{green}{\colorbox{orange!11.386000000000001}{\strut people}} \colorbox{orange!5.92}{\strut About} \mybox{blue}{\colorbox{orange!6.4030000000000005}{\strut live}} \colorbox{orange!14.869}{\strut Tokyo} \colorbox{orange!0}{\strut miilion.} &  \\
\hline
\end{tabular}

\caption{
Each RTE example contains a pair of premises and hypotheses \ie (P, H).
We shuffled the words in the original hypothesis H to create modified hypotheses (H$_{1}$ \& H$_{2}$) and fed them together with the original premise (P) to a RoBERTa-based classifier for predictions.
Also, the classifier harnessed self-attention to detect the correspondence between similar words that appear in both the premise and hypothesis.
That is, the top-3 pairs of words that were assigned the largest premise-to-hypothesis weights in a self-attention matrix (layer 0, head 7) are inside in the \textcolor{red}{red}, \textcolor{green}{green}, and \textcolor{blue}{blue} rectangles.
}
\label{fig:qualitative_examples_rte}
\end{figure*}

\clearpage
\newpage


\begin{table*}[ht]
\centering\small
\setlength\tabcolsep{5pt}
\begin{tabular}{llllrrrrr}
\toprule
& Task Name & Task Type & Label & \multicolumn{4}{c}{GLUE dev-set processing} & \devr \\
\cmidrule(l{2pt}r{2pt}){5-8}
& & & & (a) dev set & (b) step 1 & (c) step 2 & (d) step 3 & Total \\
\toprule 
& \multirow{2}{*}{\texttt{CoLA}} & \multirow{2}{*}{Acceptability} & \class{unacceptable} & 322 & 287 & 154 & 154 & 308\\
& & & \class{acceptable} & 721 & 675 & 638 & 154 & \\
\midrule
& \multirow{2}{*}{\texttt{RTE}} & \multirow{2}{*}{NLI} & \class{not entailment} & 131 & 131 & 72 & 72 & 144 \\
& & & \class{entailment} & 146 & 145 & 127 & 72 & \\
\midrule
& \multirow{2}{*}{\texttt{QQP}} & \multirow{2}{*}{Paraphrase} & \class{not duplicate} & 25,545 & 22,907 & 20,943 & 12,683 & 25,366\\
& & & \class{duplicate} & 14,885 & 14,000 & 12,683 & 12,683 & \\
\midrule
& \multirow{2}{*}{\texttt{MRPC}} & \multirow{2}{*}{Paraphrase} & \class{not equivalent} & 129 & 129 & 101 & 101 & 202\\
& & & \class{equivalent} & 279 & 279 & 255 & 101 & \\
\midrule
& \multirow{2}{*}{\texttt{SST-2}} & \multirow{2}{*}{Sentiment} & \class{negative} & 428 & 427 & 402 & 402 & 804 \\
& & & \class{positive} & 444 & 443 & 420 & 402 & \\
\midrule
& \multirow{2}{*}{\texttt{QNLI}} & \multirow{2}{*}{NLI} & \class{not entailment} & 2,761 & 2,741 & 2,500 & 2,500 & 5,000 \\
& & & \class{entailment} & 2,702 & 2,690 & 2,527 & 2,500 & \\
\midrule
& \texttt{STS-B} & Similarity & N/A & 1,500 & 1,498 & N/A & N/A & 1,498 \\
\bottomrule
\end{tabular} 

\caption{
The number of examples per class before (a) and after each of the three filtering steps to produce \devr sets (described in Sec.~\ref{sec:shuffled_sets}) for RoBERTa-based classifiers.
For each task, we repeated the same procedure for three sets of classifiers, for BERT-, RoBERTa-, ALBERT-based classifiers, respectively.
}
\label{table:preprocessing_stats}
\end{table*}

\begin{table*}[ht]
\centering\small
\setlength\tabcolsep{2.3pt}
\begin{tabular}{llcccccccccc}
\toprule
Model & Task & \devr & \devs & \multicolumn{4}{c}{\devs performance} & \multicolumn{4}{c}{Word-Order Sensitivity} \\
\cmidrule(l{2pt}r{2pt}){5-8} \cmidrule(l{2pt}r{2pt}){9-12}
&  & performance & baseline & 2-noun swap & 1-gram & 2-gram & 3-gram & 2-noun swap & 1-gram & 2-gram & 3-gram \\
\toprule
\robertaLarge & \texttt{CoLA}  & 100 & 50 & 70.80 & 51.40 & 55.62 & 57.98 & \textbf{0.58} & \textbf{0.97} & \textbf{0.89} & \textbf{0.84} \\
& \texttt{RTE}  & 100 & 50 & 82.29 & 73.85 & 80.42 & 83.75 & 0.35 & 0.52 & 0.39 & 0.33 \\
& \texttt{SST-2}  & 100 & 50 & 98.24 & 83.71 & 88.16 & 90.43 & 0.04 & 0.33 & 0.24 & 0.19 \\
& \texttt{MRPC}  & 100 & 50 & 98.54 & 85.53 & 88.64 & 90.49 & 0.03 & 0.29 & 0.23 & 0.19 \\
& \texttt{QQP}  & 100 & 50 & 87.13 & 86.84 & 90.65 & 92.60 & 0.26 & 0.26 & 0.19 & 0.15 \\
& \texttt{QNLI} & 100 & 50 & 95.26 & 91.12 & 95.20 & 96.46 & 0.09 & 0.18 & 0.10 & 0.07 \\
& \texttt{STS-B} & 90.43 & N/A & 88.95 & 85.47 & 87.20 & 87.98 & N/A & N/A & N/A & N/A \\
\bottomrule
\end{tabular}

\caption{Accuracy of all models on \devs examples (created by shuffling n-grams and swapping 2 nouns) and their Word-Order Sensitivity scores ($\in [0,1]$) across seven GLUE tasks.
STS-B is a regression task and thus not comparable in word-order sensitivity with the other tasks, which are binary classification.
}
\label{table:shuffle}
\end{table*}

\begin{table*}[t]
\centering\small
\setlength\tabcolsep{10pt}
\begin{tabular}{lccccccc}
\toprule
& \multicolumn{7}{c}{GLUE dev set} \\
\cmidrule(l{2pt}r{2pt}){2-8}
Task & CoLA & RTE & QQP & MRPC & SST-2 & QNLI & STS-B \\
& (Acc) & (Acc) & (Acc) & (Acc) & (Acc) & (Acc) & (Spearman Corr) \\
\toprule
\robertaBase & 82.56 & 72.20 & 91.12 & 87.25 & 94.50 & 92.57 & 90.17 \\
\cmidrule(l{2pt}r{2pt}){0-7}
\albertBase & 81.21 & 72.20 & 90.25 & 87.99 & 91.40 & 91.78 & 90.82 \\
\cmidrule(l{2pt}r{2pt}){0-7}
\bertBase & 81.89 & 64.25 & 90.81 & 85.54 & 92.09 & 91.38 & 88.49 \\
\cmidrule(l{2pt}r{2pt}){0-7}
\robertaLarge & 65.30 & 80.87 & 91.62 & 88.48 & 96.44 & 94.45 & 90.44 \\
\cmidrule(l{2pt}r{2pt}){0-7}
Average & 82.78 & 72.38 & 90.95 & 87.32 & 93.61 & 92.55 & 89.98 \\
\bottomrule
\end{tabular}

\caption{
The dev-set performance of models finetuned from three different BERT ``base'' variants (12 self-attention layers and 12 heads) and one RoBERTa ``large'' model (24 self-attention layers and 16 heads) on seven GLUE tasks.
These results match either those reported by original papers, \citealt{huggingface2020pretrained} or GLUE leaderboard.
}
\label{table:original_performance}
\end{table*}







\end{document}